\documentclass{article}

\usepackage{packages}

\newlength{\figfullwidth}
\newlength{\fighalfwidth}
\icmltitlerunning{Discovering User Types}

\begin{document}

\twocolumn[
\icmltitle{Discovering User Types: Mapping User Traits by Task-Specific Behaviors in Reinforcement Learning}

\icmlsetsymbol{equal}{*}

\begin{icmlauthorlist}
\icmlauthor{Lars L. Ankile}{equal,seas}
\icmlauthor{Brian S. Ham}{equal,seas}
\icmlauthor{Kevin Mao}{seas}
\icmlauthor{Eura Shin}{seas}
\icmlauthor{Siddharth Swaroop}{seas}
\icmlauthor{Finale Doshi-Velez}{seas}
\icmlauthor{Weiwei Pan}{seas}
\end{icmlauthorlist}

\icmlaffiliation{seas}{Harvard University, Cambridge MA, USA}

\icmlcorrespondingauthor{Lars L. Ankile}{larsankile@g.harvard.edu}

\icmlkeywords{Machine Learning, ICML, mHealth, Reinforcement Learning}

\vskip 0.3in
]

\printAffiliationsAndNotice{\icmlEqualContribution}

\begin{abstract}
    When assisting human users in reinforcement learning (RL), we can represent users as RL agents and study key parameters, called \emph{user traits}, to inform intervention design.
    We study the relationship between \emph{user behaviors} (policy classes) and user traits. 
    Given an environment, we introduce an intuitive tool for studying the breakdown of ``user types": broad sets of traits that result in the same behavior.
    We show that seemingly different real-world environments admit the same set of user types and formalize this observation as an equivalence relation
    defined on environments. 
    By transferring intervention design between environments within the same equivalence class, we can help rapidly personalize interventions.
\end{abstract}

\section{Introduction}
\label{introduction}

Mobile Health (mHealth) applications, like a physical therapy (PT) app that recommends personalized exercises to a user working to regain ankle mobility, are gaining popularity as cost-effective interventions. In these applications, personalization can be achieved by inferring the user-specific internal obstacles to reaching health targets, then designing treatments for those obstacles \cite{shin2022modeling}. In this paper, we provide a set of novel tools for studying the relationship between user-specific obstacles and user behavior, thereby generating insights for treatment design.

We model user-internal obstacles as discrepancies between the real-world environment, formalized as a Markov Decision Process (MDP), and the user's perceived environment, another MDP. We want the user, as a Reinforcement Learning (RL) agent, to adopt a target policy in the real-world environment (e.g., perform recommended daily exercises until full recovery), but since the user plans in their perceived environment, their perceived optimal policy can deviate drastically from the target (e.g., prematurely terminate the PT program due to the perceived infeasibility of full recovery). 

For a real environment, we characterize the user-perceived environments using MDP parameters that map to well-studied human traits---which we call \emph{user traits}---in the behavioral sciences. In particular, for many mHealth applications, a user's confidence in their physical capabilities and their ability to perform long-term planning (their degree of myopia) both significantly impact their success in prescribed fitness regimens \citep{picha2021physicalTherapySelfEfficacy}. In our work, we model \emph{myopia} as the discount factor and \emph{confidence} as the dynamics (specifically, the perceived probability of positive outcomes) of the user's MDP (\autoref{sec:users_as_rl_agents}).

Given a real environment, we introduce a tool for visualizing the relationship between user traits (the user's MDP parameters) and the corresponding user behavior (the user's possible policies). Specifically, within the environment, we study the breakdown of ``user types"---regions in the space of all possible user traits that define the same user behavior---and visualize these types as two-dimensional \textit{behavior maps} (\autoref{sec: behavior maps}). Behavior maps shed light on the extent to which it is possible to infer user traits by observing user behavior.

Finally, we show that seemingly different real-world environments admit the same behavior maps. We formalize this observation as an equivalence relation defined on real-world environments (\cref{sec: equivalence relation}). We map several environments commonly used in the RL literature (that also model mHealth tasks) to just a small set of equivalence classes, where the sets of user behaviors are similar across different environments within each class (\cref{sec: rich classes}). This result allows us to provide guidelines on intervention design in various complex environments by lifting insights from an equivalent and simpler toy environment (\cref{sec: equivalence design}).

\section{Related Work}

\paragraph{Inferring a user's parameters from demonstrations.} 
Like us, some works \citep{evans2016learningIgnorant,pmlrv97shah19FeasibilityOf, zhi2020onlineGoalInference} model humans as RL agents with different perceived MDPs. However, inferring an agent's MDP parameters from demonstration is a difficult and nonidentifiable problem \citep{pmlrv97shah19FeasibilityOf}.
This paper shows that, while user parameters cannot be exactly recovered from behavior data in most settings, we can infer general rules about the relationship between user parameters and user behavior. These rules can help us design mHealth interventions.

\paragraph{Equivalence in Inverse RL (IRL).} 
In IRL, when parameters of an MDP cannot be uniquely identified, we infer classes of these parameters, typically rewards \citep{ziebart2010MaxCausalIRL} or transitions functions \citep{reddy2018inverseTransitions,golub2013learningInternalDynamicsControl}, that are equally likely under the behavior data provided by \emph{one} user. 
In this work, we study the behaviors of \emph{multiple users} and equate different environments (MDPs) in which the partitioning of the set of users by behavior is similar.

\paragraph{Equivalence of MDPs.} 
Notions of equivalence between MDPs allow for knowledge transfer between different environments \citep{soni2006usingHomoTransfer,sorg2009transferHomo}. 
For example, bisimulation-based equivalence definitions are used in MDP minimization, where large state spaces are reduced to speed up planning \citep{givan2003equivalence}.
Relaxed versions of bisimulations, e.g., MDP homomorphism \citep{biza2018onlineAbstractMDPHomo}, stochastic homomorphism \citep{van2020plannableHomo}, and approximate homomorphisms \citep{ravindran2004approximateHomo} allow optimal policies in simple MDPs to be lifted to desirable policies in more complex and comparable MDPs.
More general definitions of MDP equivalence can be defined through other methods of state aggregation (e.g., value equivalence) \cite{li2006towards}.
While these notions of equivalence are defined over the set of MDPs,  we decompose an MDP into task-specific and user-specific components and consider equivalences between the task-specific components of MDPs while varying the user-specific ones.

\section{Formalizing Users as RL Agents}
\label{sec:users_as_rl_agents}

We formalize an RL environment for an mHealth application as a Markov Decision Process (MDP). An MDP is a 5-tuple, $\mathcal{M} = \langle  \mathcal{S}, \mathcal{A}, T, R, \gamma \rangle$, consisting of a set of states $\mathcal{S}$, a set of actions $\mathcal{A}$, a reward function $R: \mathcal{S} \times \mathcal{A} \times \mathcal{S} \to \mathbb{R}$, a transition function $T: \mathcal{S} \times \mathcal{A} \times \mathcal{S} \to [0, 1]$ and a discount rate $\gamma \in [0, 1]$. For simplicity, in this paper, we only consider discrete state spaces. 

An optimal RL agent acts in $\mathcal{M}$ according to a policy $\pi_\mathcal{M}: \mathcal{S} \rightarrow \mathcal{A}$, giving a cumulative reward (expected returns): $J^\pi_\mathcal{M} = \mathbb{E}\left[\sum\limits_{t=0}^T \gamma^t r_t \right]$, where $r_t$ is the random variable representing the reward received at time $t$. The optimal policy for $\mathcal{M}$ maximizes the expected returns: $\pi^*_\mathcal{M}  = \max\limits_{\pi}J^\pi$.

We want the user to adopt the optimal policy $\pi^*_\mathcal{M}$ in $\mathcal{M}$. However, the user plans in their perceived environment, $\mathcal{M}^{\text{user}} = \langle  \mathcal{S}^\text{user}, \mathcal{A}^\text{user}, T^\text{user}, R^\text{user}, \gamma^\text{user} \rangle$ and adopts the policy, $\pi^*_{\mathcal{M}^\text{user}}$, that is optimal for $\mathcal{M}^\text{user}$. Discrepancies between the real environment and the user's perceived one can lead to drastic differences between the target policy, $\pi^*_\mathcal{M}$, and the adopted one, $\pi^*_{\mathcal{M}^\text{user}}$.

In this work, we shall assume that the perceived environment differs from the real only in the transition function (modeling the user trait confidence) and the discount rate (modeling the user trait myopia). Specifically, we define a \textbf{\emph{world}} as a tuple $\mathcal{W} = \langle\mathcal{S}, \mathcal{A}, R\rangle$ of states $\mathcal{S}$, actions $\mathcal{A}$, and reward function $R$. This captures the real environment and the task in an application of interest (see \cref{fig:behaviors_examples} for example-grid worlds). Since the user's perceived states, actions, and rewards match the real environment, we set $\mathcal{S}^\text{user}=\mathcal{S}$, $\mathcal{A}^\text{user} = \mathcal{A}$ and $R^\text{user} = R$.

Furthermore, since we are interested in the set of optimal policies generated by varying the user's perceived environment $\mathcal{M}^\text{user}$, we do not keep track of the real transition function $T$ and the real discount rate $\gamma$. Instead, the user's policy depends only on their (fixed) perception of the environment, $T^\text{user}$, and their (fixed) discount rate $\gamma^\text{user}$. The real $T$ is useful only when the user is learning (updating $T^\text{user}$) based on data generated by $T$.

We use $\gamma^\text{user}\in [0, 1]$ to represent the user's level of \emph{myopia}. To represent the level of \emph{confidence}, we parameterize the user's transition $T^\text{user}_p$ function with $p \in [0, 1]$, which is the level of stochasticity in the environment transitions that the user perceives. Other parameterizations of $T^\text{user}$ are possible, but this one aligns with the intuition that a user with low confidence is unsure whether their actions $a \in \mathcal{A}^\text{user}$ will lead to desired outcomes $s' \in \mathcal{S}^\text{user}$.

In \cref{sec: behavior maps}, we model how and why users with distinct traits behave differently (i.e., adopt different policies) in the same real-life setting. 
For example, two people with different levels of myopia would judge different PT behaviors to be optimal in their respective MDPs.
However, we first connect our formalization of user traits (their level of myopia $\gamma^{\text{user}}$, and their confidence level $p$ parameterizing $T^\text{user}_p$) to well-studied constructs in psychology and behavioral science. 

\paragraph{Mapping RL to Behavior Science.}
\emph{Myopia} corresponds to the concept of temporal discounting in psychology. In user MDPs, we represent temporal discounting with $\gamma^\text{user} \in [0, 1)$. This captures people's tendency to undervalue future rewards, often leading to unhealthy behavior \citep{story2014does}. However, we note that in RL, discounting is exponential by default, which does not capture the phenomenon observed in humans called \textit{preference reversal} \cite{ainslie1992hyperbolic,pmlrv97shah19FeasibilityOf} (which \textit{hyperbolic discounting} is more suited for). 

In behavioral science, \emph{confidence}, also known as self-efficacy, measures an agent's belief in their capability to perform a task \cite{picha2021physical}. Intuitively, this is the user's perceived probability that their intended outcome can be achieved through action. In user MDPs, we represent the user's confidence level with $p \in [0, 1]$, which is the level of stochasticity in the transitions. Concretely,  $T^\text{user}_p(s, a, s') = p$ for a user's intended outcome $s'$ from performing action $a$ in state $s$. We divide the remaining $1-p$ probability equally among the alternate outcomes: $T^\text{user}(s, a, \hat s') = \frac{1-p}{|\hat{\mathcal{S}}|}$. Our current instantiation of confidence is simple, and it is equivalent to adding epsilon-noise to the real-world transition matrix. However, the transition $T^\text{user}_p$ can be a function of $p$ in more complex ways.

\begin{figure}
    \centering
    \includegraphics[width=0.8\linewidth]{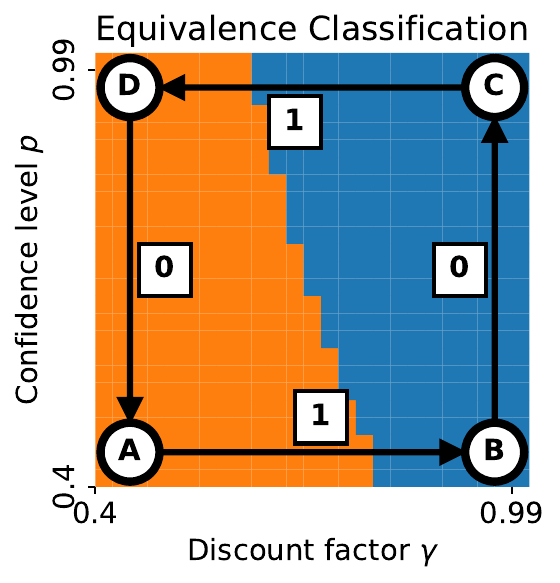}
    \caption{Example behavior map (Big-Small world). 
    The two colors indicate the two possible behaviors (see \cref{fig:behaviors_big_small} for the world and the behaviors). 
    Annotations describe the procedure for deriving the equivalence class. The x-axis varies over the discounting factor, $\gamma$; the y-axis varies over the confidence level, $p$. 
    ``Extreme" users, i.e., corners of the map, are labeled as circles. The number of ``behavior switches" when tracing each edge between extreme users (from A to B, to C, to D, and back to A) are labeled as squares.
    }
    \label{fig:behaviormap_sketch}
\end{figure}

\section{Behavior Maps: A Tool for Understanding User Traits and User Behaviors}
\label{sec: behavior maps}

In the previous section, we formalized the user's MDP $\mathcal{M}^{\text{user}}$ and their optimal policy $\pi^*_{\text{user}}$. 
We now introduce behavior maps, a tool for studying the relationship between the user-specific parameters ($T^{\text{user}}_p$, $\gamma^{\text{user}}$) and the corresponding optimal user policy $\pi^*_{\text{user}}$.

Given a world $\mathcal{W}$, we denote the set of possible (deterministic) policies, $\pi: \mathcal{S} \to \mathcal{A}$, as $\Pi_\mathcal{W}$. 
We note that in many real-life applications, distinct policies may functionally describe the same type of behavior (e.g., if we are interested in overall adherence, skipping PT exercises every Tuesday can be considered functionally equivalent to skipping every Monday). 
Thus, we work with a concept that generalizes the notion of policy; we define a ``\emph{user behavior}", denoted $B \subset \Pi_\mathcal{W}$, as a set of policies considered \emph{equivalent} in the application domain. 
We study how differences in user traits lead to different user behaviors. 

To do this, we introduce a \textbf{\emph{behavior map}} of the world $\mathcal{W}$ as a mapping of user traits to the corresponding user behaviors in $\mathcal{W}$. 
That is, the behavior map $\mathcal{B}_{\mathcal{W}}$ maps $(\gamma^{\text{user}}, p)$ to the user behavior $B$ that contains the optimal policy for the user MDP $\mathcal{M}_{\text{user}} = \langle \mathcal{S}, \mathcal{A}, \mathcal{R}, T^\text{user}_p, \gamma^{\text{user}}\rangle$. 

In \cref{fig:behaviormap_sketch}, we show an example of a behavior map. 
We see that it classifies the user parameter space 
into regions where parameters map to the same user behavior. 
In this world, there are only two behaviors (indicated by color), and the user's behavior depends on the value of their user traits (the two axes).

\paragraph{Applications of Behavior Maps.}
We demonstrate that behavior maps can inform the design and deployment of interventions on user traits (for example, interventions to increase $\gamma^{\text{user}}$). Specifically, they can help us (1) determine to what extent user traits are identifiable through behavioral observations; (2) warm-start an intervention strategy for interacting with new users.

\paragraph{Identifiability of User Traits.}
Since behavior maps tell us which set of parameters gives the same user behavior, they allow us to anticipate the limits of what we can infer about a user (using Inverse Reinforcement Learning (IRL) or related methods) by observing their behavior in a given world. 
For example, in worlds with the behavior map in \cref{fig:behaviormap_sketch}, we can distinguish between users with low and high discount factors because users have different optimal policies (different colors). 
On the other hand, the difference in confidence does not generally correspond to a difference in behavior. Therefore, we cannot generally distinguish between users with different confidence levels.
However, we find that behavior maps can inform intervention design, even when the parameters of individual users cannot be exactly inferred.
 
\paragraph{Warm-start Intervention Strategy.}
Given a world and a new user, behavior maps can help identify interventions that, a priori, is likely to be more impactful. In particular, the more variation there is in user behavior along a given axis, the more likely an intervention on the corresponding trait will change the user's behavior. For example, in \cref{fig:behaviormap_sketch}, we know that an intervention on $\gamma^{\text{user}}$ is more likely to change the user's behavior than an intervention on $T_p^{\text{user}}$.

Although useful, directly computing the behavior map for a complex application such as PT requires solving user MDPs for a range of user parameters and can thus be computationally costly. Instead, to get the same insights, we reduce the PT world $\mathcal{W}$ to a simpler toy world $\mathcal{W}'$, for which we can easily compute $\mathcal{B}_\mathcal{W'}$. We define an equivalence relation that allows us to make this reduction.

\section{A Behavior-Based Equivalence Relation}
\label{sec: equivalence relation}
This section uses behavior maps to draw analogies between seemingly different worlds. 

Suppose that two different applications, such as PT and dieting, have the same behavior map, such as the one from \cref{fig:behaviormap_sketch}. Then, in both applications, we know that confidence does not impact user behavior and that users with ``low'' gamma have one behavior, while users with ``high'' gamma have another. In this way, we consider PT and dieting equivalent worlds because intervention design principles can be transferred from one to the other. For example, in both cases, the initial intervention strategy should focus on $\gamma^{\text{user}}$ instead of $T_p^{\text{user}}$. Note that this transfer can work in cases where the state and action spaces differ between the two applications because the behavior maps depend on \emph{high-level behaviors} (not exact states and actions). For example, in PT, the behaviors may be a set of exercises. In dieting, the behaviors may be a set of food choices. In either case, there is a desired behavior (e.g., choosing nutritious foods or choosing the right exercises) and an undesired behavior. We are only concerned with what interventions will help the user go from undesired to desired behaviors, not that the actions defining those behaviors match exactly.

Moreover, we can transfer between worlds with similar but not necessarily identical behavior maps.
For example, we might see that both PT and dieting have two possible behaviors, where users with lower $\gamma^{\text{user}}$ act differently from users with higher $\gamma^{\text{user}}$. However, what is considered to be ``low" or ``high" $\gamma^{\text{user}}$ need not match exactly between the two applications: in PT, the range for ``low" $\gamma^{\text{user}}$ could be $[0, 0.3]$ and in dieting the range could be $[0, 0.2]$. If we knew both applications had similar behavior maps, we could still transfer the knowledge that the initial intervention strategy should focus on $\gamma^{\text{user}}$ instead of $T_p^{\text{user}}$. We could also transfer the knowledge that users with different $\gamma^{\text{user}}$ are identifiable, while users with different $T_p^{\text{user}}$ are not.

\subsection{Equivalence Between Behavior Maps}

Thus motivated, we call two behavior maps equivalent if the \emph{shapes} of the decision boundaries between user behaviors in the behavior maps are the same and use an equivalence definition invariant to stretching or translation of these boundaries. We formalize this in \cref{def:equiv}.
 
\begin{figure*}[t]
    \centering
    \begin{subfigure}{0.32\linewidth}
         \centering
         \includegraphics[width=1\linewidth]{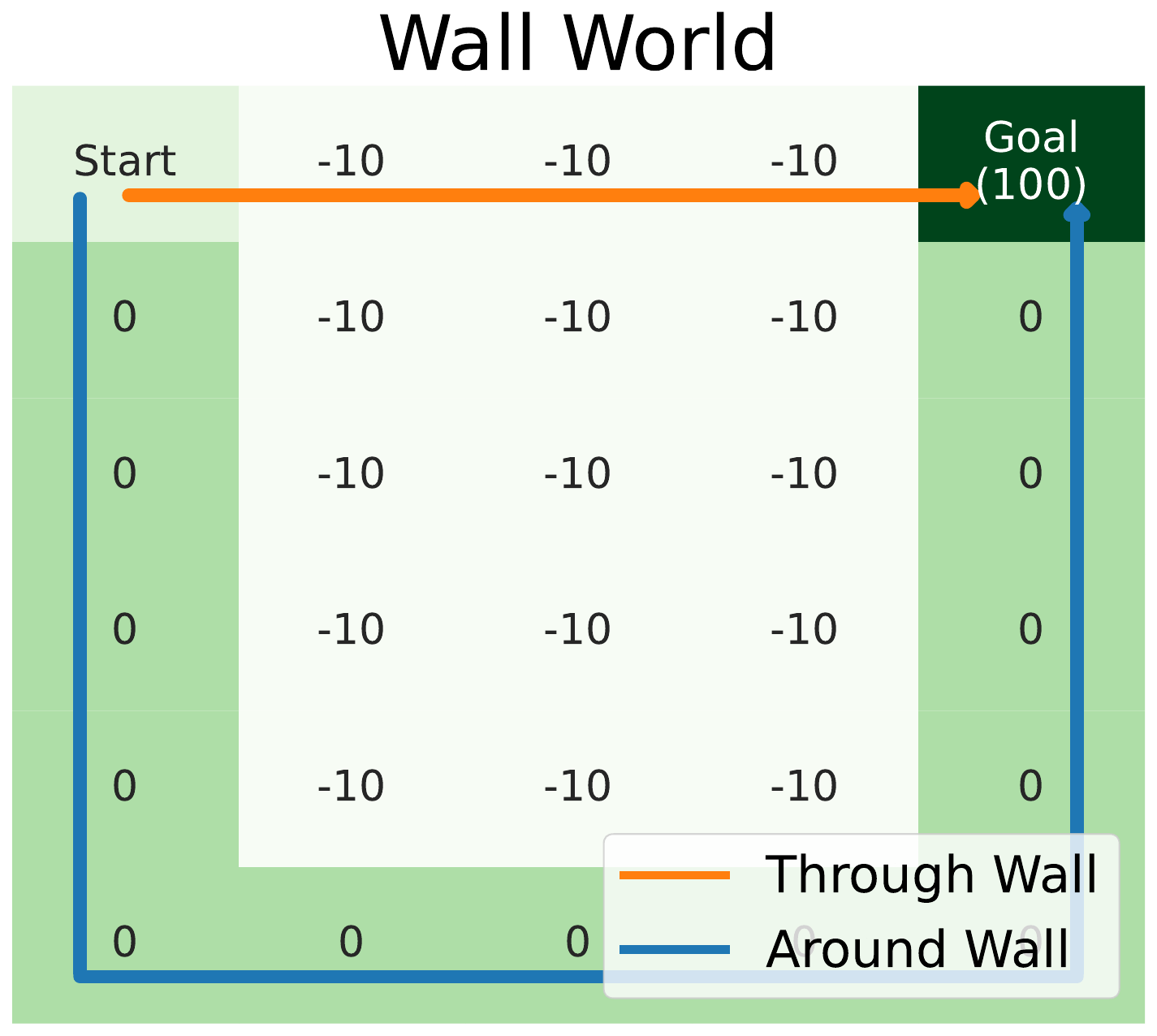}
         \caption{A $6\times 5$ \textit{Wall world} where agents can pass directly through a costly wall (orange) or take the longer, safer path around it (blue).}
         \label{fig:behaviors_wall}
     \end{subfigure}\hfill
    \begin{subfigure}{0.32\linewidth}
         \centering
         \includegraphics[width=1\linewidth]{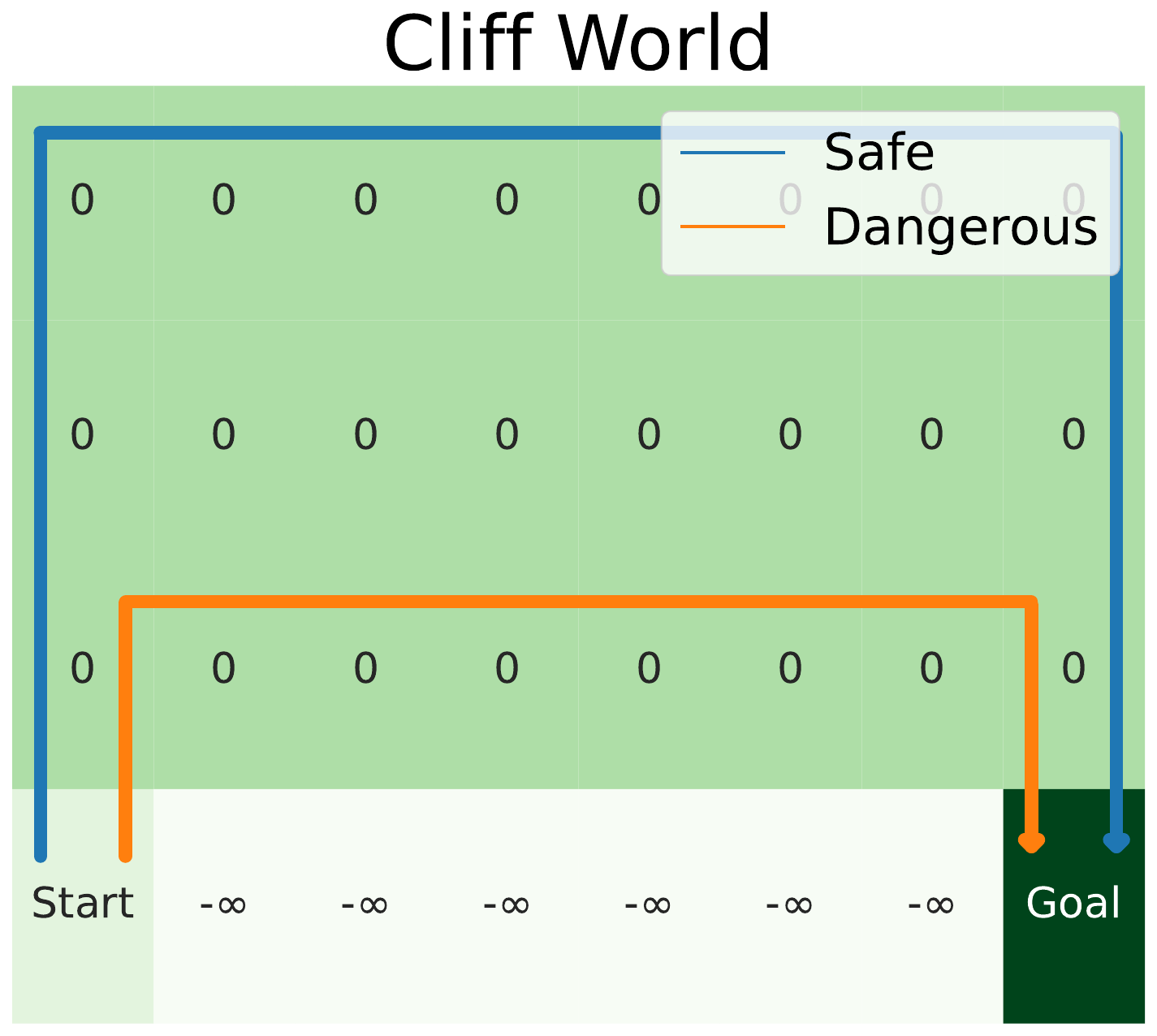}
         \caption{A $4\times 8$ \textit{Cliff world} where agents can walk close to the cliff and risk ruin (blue) or keep space but walk farther (orange).}
         \label{fig:behaviors_cliff}
    \end{subfigure}\hfill
    \begin{subfigure}{0.32\linewidth}
         \centering
         \includegraphics[width=1\linewidth]{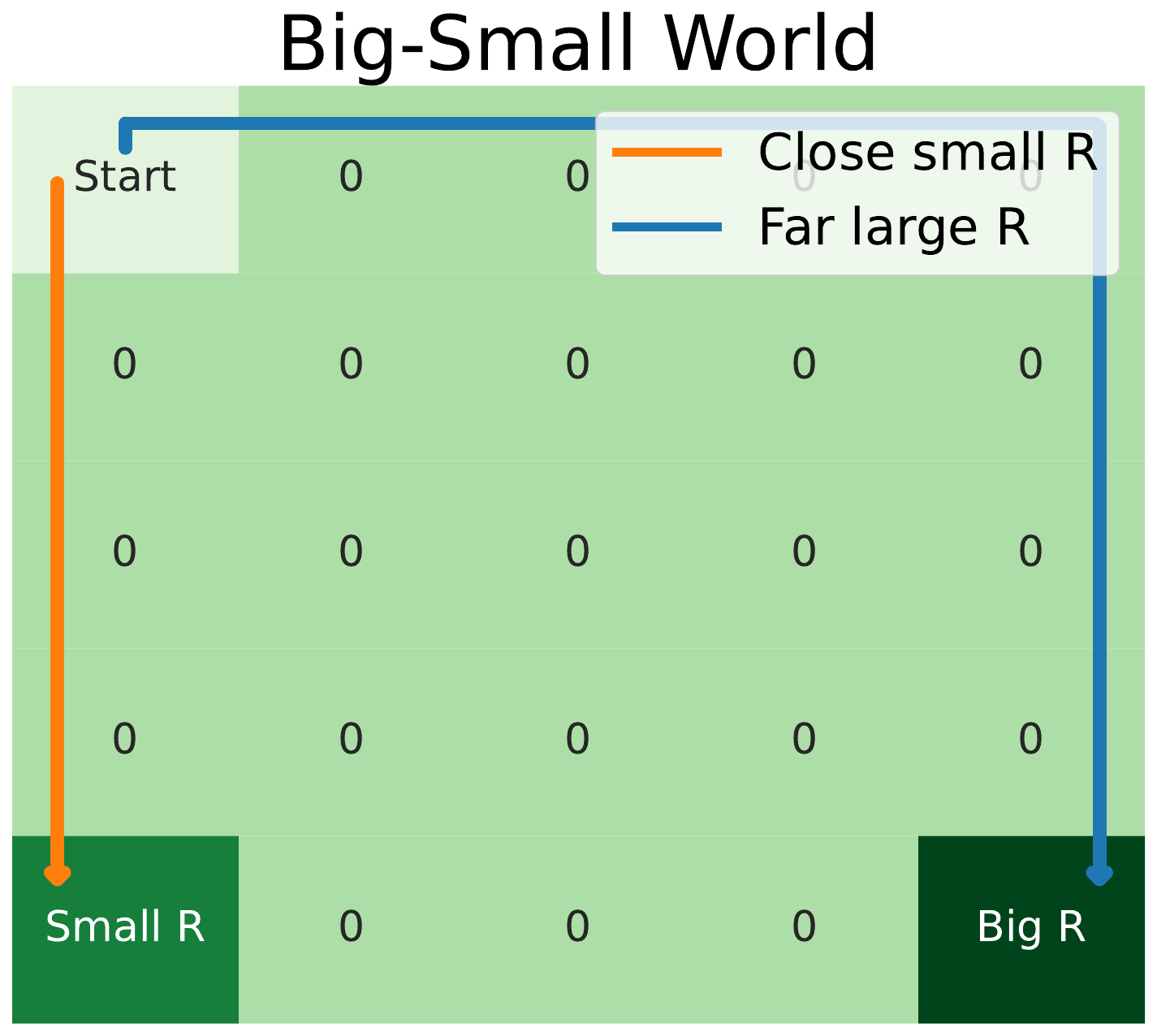}
         \caption{A $5\times 5$ \textit{Big-Small world} where agents can walk straight down to a small reward (orange) or farther to a bigger reward (blue).}
         \label{fig:behaviors_big_small}
    \end{subfigure}
    \caption{Each atomic world has two qualitatively distinct behaviors (shown with blue and orange arrows). Each diagram shows what the world looks like for one setting of the parameters, and other sizes are usually also valid.}
    \label{fig:behaviors_examples}
\end{figure*}

In the following, we assume, without loss of generality, that the axes of each behavior map $\mathcal{B}_{\mathcal{W}}$ is scaled to the unit interval, that is, $\mathcal{B}_{\mathcal{W}}$ is a map over $I^2$, where $I = [0,1]$. Thus, the decision boundary classifying different user behaviors in $\mathcal{B}_{\mathcal{W}}$ is a 1-dimensional submanifold in $I^2$ defined by the map $g_\mathcal{W}: [0, 1] \to I^2$ satisfying some additional constraints. Although we consider the case where the decision boundary is connected here, our definition extends straightforwardly to cases where it is not.

\begin{defn}[World Equivalence Induced by Behavior Map]
\label{def:equiv}
We define an equivalence relation, $\equiv_{\mathrm{map}}$, on the set of discrete worlds $\mathfrak{W}$ by
$$
\mathcal{W} \equiv_{\mathrm{map}} \mathcal{W}', \quad \mathcal{W}, \mathcal{W}' \in \mathfrak{W}
$$
when (1) the number of behaviors in $\mathcal{B}_{\mathcal{W}}$ and $\mathcal{B}_{\mathcal{W}'}$ are equal, and (2) there is
a continuous map $h: I^2 \times [0,1] \to I^2$, such that $h_t: I^2 \times \{t\} \to I^2$ is bijective, where $h_0$ is the identity map, and where $h_1$ satisfies $h_1 \circ g_\mathcal{W} = g_{\mathcal{W}'}$.
\end{defn}

Note that we can simply say that $h$ is an \emph{ambient isotopy} between the decision boundaries in $\mathcal{W}$ and $\mathcal{W}'$.

The idea behind \cref{def:equiv} can be made more intuitive. 
We consider each behavior map as a diagram in which (i) $n_i$ number of vertices (each representing a switch between behaviors) is placed on the $i$-th edge, and where (ii) each pair of vertices is connected by a curve defined by a decision boundary that separates two user behaviors (see \cref{fig:behaviormap_sketch}). 
We say that two maps are equivalent if they are labeled by the same number of distinct behaviors, and, as diagrams, they are topologically equivalent: the decision boundary in one behavior map can be continuously deformed, by using the map $h$, to look like that in the other. 

For the set of worlds studied in this work, we note that whether two worlds are equivalent boils down to counting the number of behavior switches along the edges of their behavior maps (counterclockwise, starting from the bottom edge). We can focus exclusively on the edges because our worlds do not induce behavior maps with decision boundaries that behave differently in the middle (e.g. \cref{fig:behavior_map_middle_2}). By counting the number of behavior switches along the edges, we can represent the set of worlds in the same equivalence class as a count vector (see \cref{fig:behaviormap_sketch}).

\subsection{Intervention Transfer Between Equivalent Worlds}
\label{sec: transfer}

Recall that our primary motivation for defining an equivalence relation on worlds is to develop intervention strategies in simple settings and transfer them to more complex analogous ones. This section provides the formalism for transferring interventions between equivalent worlds. In \autoref{sec: atomic}, we will introduce a set of simple worlds to which many commonly studied RL environments can be reduced through our equivalence.

Given a world $\mathcal{W}$, we represent a single \emph{intervention} on a user's myopia and confidence level as a real-valued pair $(\Delta_\gamma, \Delta_p) \in I^2$ that is added to the user's current parameters. Thus, a sequence of interventions defines a (piece-wise linear) path, which we call an \emph{intervention strategy} and denote by $\tau_\mathcal{W}$, in the behavior map $\mathcal{B}_\mathcal{W}$. Our goal is to map an intervention strategy $\tau_{\mathcal{W}}$ in $\mathcal{B}_\mathcal{W}$, that realizes a behavior change, to a strategy $\tau_{\mathcal{W}'}$ in an equivalent map $\mathcal{B}_{\mathcal{W}'}$ that realizes an analogous behavior change. 

We first observe that the continuous map $h$ in \cref{def:equiv} induces a mapping from the set of user parameters related to one world $\mathcal{W}$ to the user parameters related to $\mathcal{W}'$, defined by 
$
h_1: I^2 \to I^2.
$
Hence, every path $\tau_\mathcal{W}$ defines a path $\tau_{\mathcal{W}'} = h_1 \circ \tau_\mathcal{W}$. 
Since $h$ continuously deforms the decision boundary of $\mathcal{B}_\mathcal{W}$, it preserves the number of times $\tau_\mathcal{W}$ intersects the decision boundary in $\mathcal{B}_\mathcal{W}$. In particular, if $\tau_\mathcal{W}$ represents an intervention strategy that achieves $N$ number of behavior changes in $\mathcal{B}_\mathcal{W}$, then $\tau_{\mathcal{W}'}$ is a strategy that achieves the same number of behavior changes in $\mathcal{B}_{\mathcal{W}'}$.

Note that, unlike knowledge generalization approaches in RL wherein one computes a mapping between all parameters of two MDPs, our approach to intervention transfer between two worlds by-passes explicit mappings between the state and action sets of $\mathcal{W}$ and $\mathcal{W}'$. Instead, we rely on $h$, the mapping between user parameter and policy spaces.

In practice, explicitly computing $h$ can be difficult. In the next section, we show that we can derive a more general set of heuristics for intervention design in a complex world by reasoning about an equivalent simple world.

\begin{figure*}[t]
    \centering
     \begin{subfigure}[t]{0.3\linewidth}
         \centering
         \includegraphics[width=0.8\linewidth]{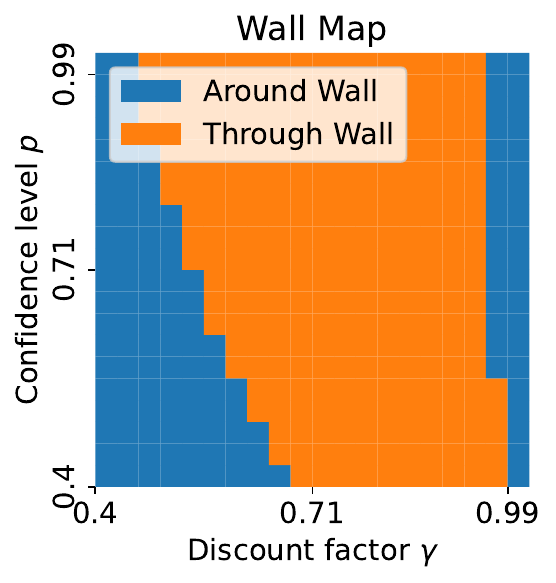}
         \caption{\emph{Wall world} example behavior map belonging to equivalence class [2, 0, 2, 0].}
         \label{fig:wall_behavior_map}
    \end{subfigure}\hfill
     \begin{subfigure}[t]{0.3\linewidth}
         \centering
         \includegraphics[width=0.8\linewidth]{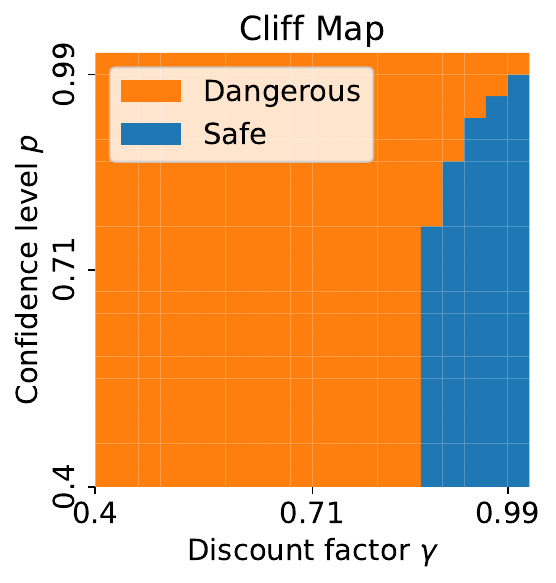}
         \caption{\emph{Cliff world} example behavior map belonging to equivalence class [1, 1, 0, 0].}
         \label{fig:cliff_behavior_map}
    \end{subfigure}\hfill
     \begin{subfigure}[t]{0.3\linewidth}
         \centering
         \includegraphics[width=0.8\linewidth]{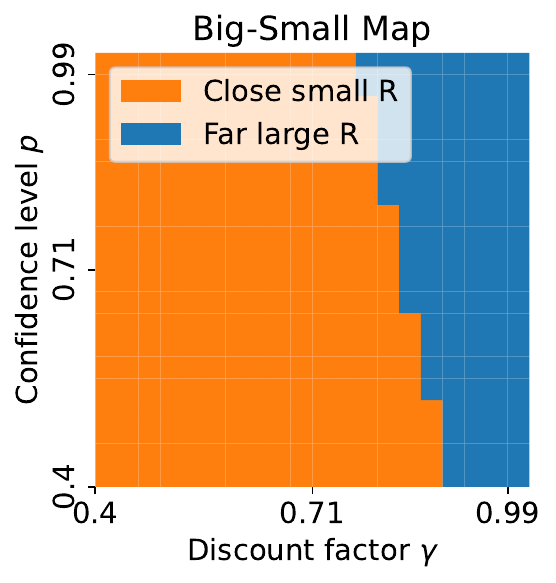}
         \caption{\emph{Big-Small} world example behavior map belonging to equivalence class [1, 0, 1, 0].}
         \label{fig:big_small_behavior_map}
    \end{subfigure}\\
    \begin{subfigure}[t]{0.18\linewidth}
         \centering
         \includegraphics[width=1\linewidth]{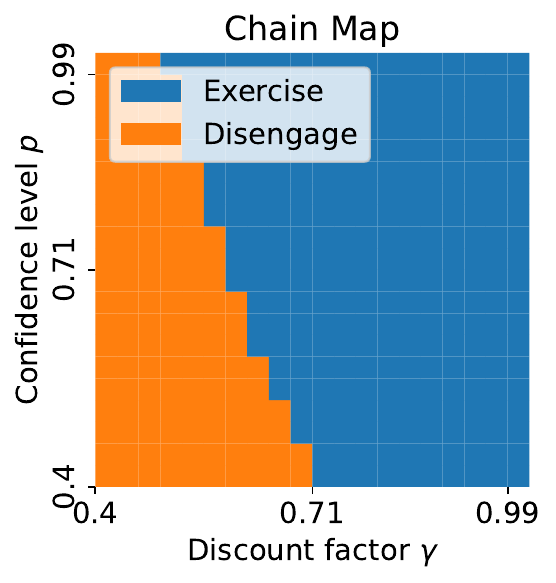}
         \caption{\emph{Chain world} example behavior map belonging to equivalence class [1, 0, 1, 0].}
         \label{fig:chain_behavior_map}
    \end{subfigure}\hfill 
    \begin{subfigure}[t]{0.18\linewidth}
         \centering
         \includegraphics[width=1\linewidth]{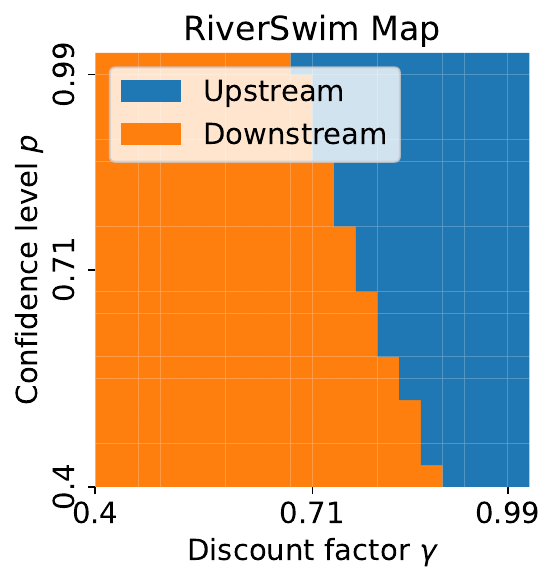}
         \caption{\emph{RiverSwim world} example behavior map belonging to equivalence class [1, 0, 1, 0].}
         \label{fig:riverswim_behavior_map}
    \end{subfigure}\hfill
     \begin{subfigure}[t]{0.18\linewidth}
         \centering
         \includegraphics[width=1\linewidth]{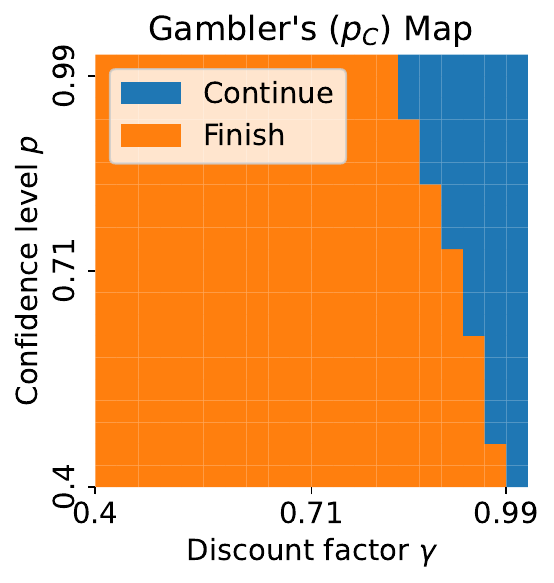}
         \caption{\emph{Gambler's Ruin world} (varying $p_C$) example behavior map in equivalence class [1, 0, 1, 0].}
         \label{fig:gambler_fc_behavior_map}
    \end{subfigure}\hfill
     \begin{subfigure}[t]{0.18\linewidth}
         \centering
         \includegraphics[width=1\linewidth]{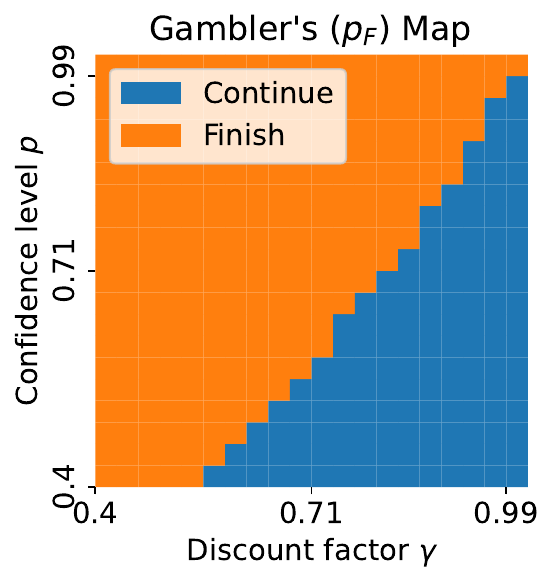}
         \caption{\emph{Gambler's Ruin world} (varying $p_F$) example behavior map in equivalence class [1, 1, 0, 0].}
         \label{fig:gambler_ff_behavior_map}
    \end{subfigure}\hfill
     \begin{subfigure}[t]{0.18\linewidth}
         \centering
         \includegraphics[width=1\linewidth]{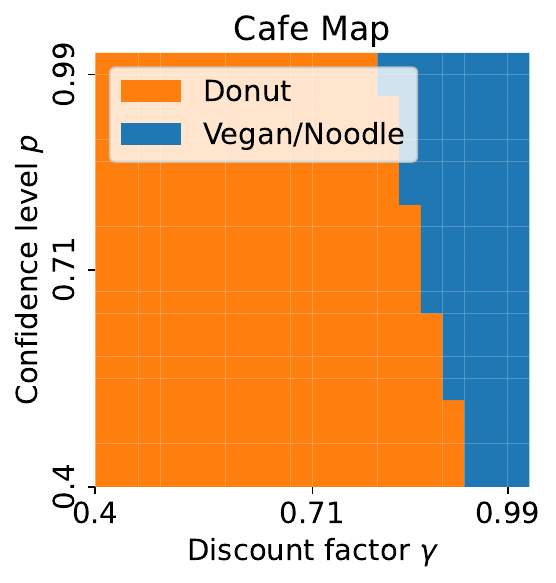}
         \caption{\emph{Café world} example behavior map belonging to equivalence class [1, 0, 1, 0].}
         \label{fig:cafe_behavior_map}
    \end{subfigure}
    \caption{Seemingly different worlds (bottom row) are equivalent to one of our atomic worlds (top row).
    }
    \label{fig:behavior_maps}
\end{figure*}

\begin{figure}[ht]
    \centering
    \begin{subfigure}[b]{0.48\linewidth}
         \centering
         \includegraphics[width=1\linewidth]{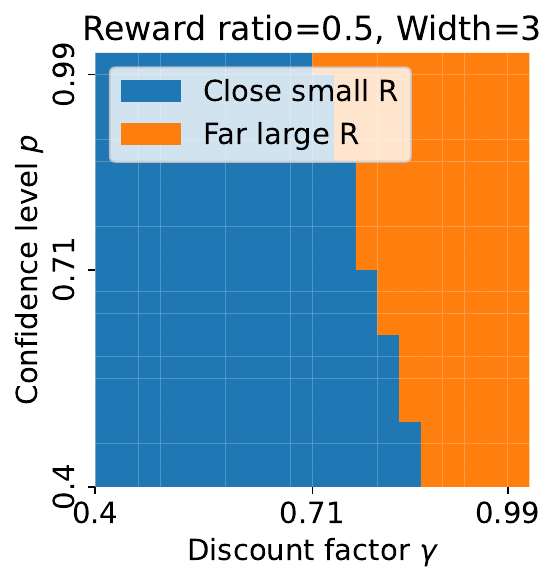}
         \label{fig:big_small_class_invariance_1}
     \end{subfigure}
    \begin{subfigure}[b]{0.48\linewidth}
         \centering
         \includegraphics[width=1\linewidth]{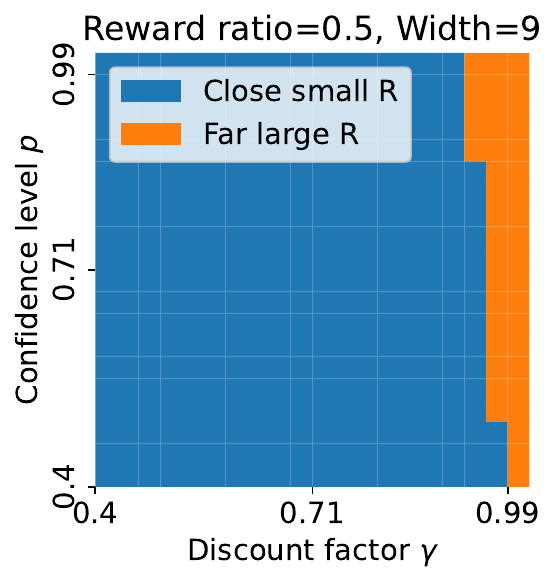}
         \label{fig:big_small_class_invariance_2}
    \end{subfigure}\\
    \begin{subfigure}[b]{0.48\linewidth}
         \centering
         \includegraphics[width=1\linewidth]{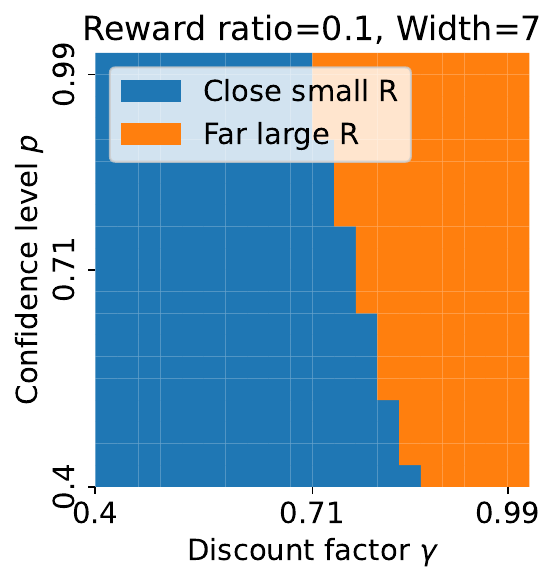}
         \label{fig:big_small_class_invariance_3}
    \end{subfigure}
    \begin{subfigure}[b]{0.48\linewidth}
         \centering
         \includegraphics[width=1\linewidth]{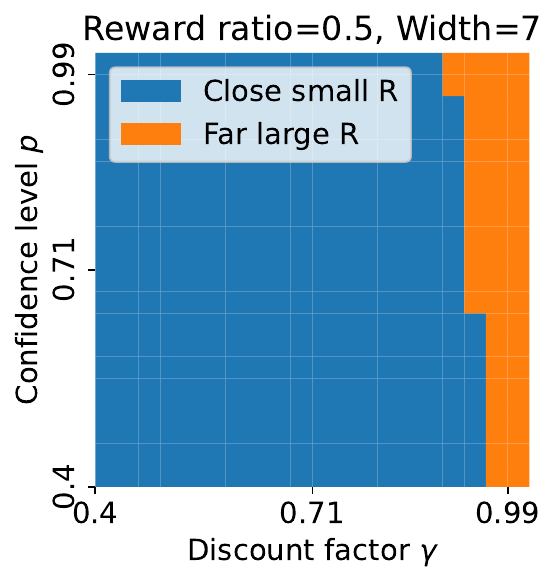}
         \label{fig:big_small_class_invariance_4}
    \end{subfigure}
    \caption{A \emph{Big-Small world} stays within its equivalence class for many different parameter combinations. The example behavior maps have different values for the world width and the ratio of the small reward to the big reward, while the rest of the parameters are fixed as \texttt{height = 7} and \texttt{Big far R = 300}.}
    \label{fig:inclass_invariance}
\end{figure}

\section{Atomic Worlds: Simple Representatives of Equivalence Classes}

Under \cref{def:equiv}, we seek the simplest representative, called \emph{atomic worlds}, for each equivalence class. User behaviors can be characterized in atomic worlds, and the insights transferred thereafter to more complex equivalent worlds. We describe three atomic worlds and reduce commonly studied worlds in RL literature to our atomic worlds. 

\subsection{Atomic Worlds}
\label{sec: atomic}
We visualize an instance of each of the following worlds in \cref{fig:behaviors_examples} and their corresponding behavior maps in \cref{fig:behavior_maps}.

The \emph{Big-Small world} is an atomic world that captures a trade-off between choosing a smaller, more convenient reward and a bigger reward that is more difficult to reach. In mHealth, this world reflects scenarios in which smaller immediate rewards, such as the time saved by skipping PT for the day, preclude larger but delayed rewards, such as a fully rehabilitated ankle.

The \emph{Cliff world} captures settings in which a harmful absorbing state may be reached due to an action going awry. For example, deciding the intensity of the PT regimen can be modeled as a Cliff world. A high-intensity regimen could accelerate recovery but also risk re-injuring the patient.

The \emph{Wall world} captures the choice between a short, costly path to the goal and a longer, free path to the same goal. This can model the trade-off in choosing the type of physical therapy: virtual therapy may be more affordable, while in-person therapy is more costly and targeted.

In the above, we note that different aspects of user decision-making (e.g., choosing the intensity vs choosing the type of therapy) in the same mHealth application (PT), can map to different equivalence classes. 
We hypothesize that more complex worlds (e.g., larger portions of the user decision-making process in PT) can be captured by compositions of simpler atomic worlds. In future work, we are interested in characterizing the set of complex worlds that can be studied through decomposition into atomic worlds. Further discussion can be found in \autoref{sec:discussion} and \autoref{app:world_composition}.

\subsection{Atomic Worlds Capture Commonly Studied RL Environments}
\label{sec: rich classes}

We compare the behavior maps corresponding to four types of RL environments commonly studied in the literature: Chain, RiverSwim, Gambler's Fallacy, and Café worlds (details on each world are in \autoref{appendix: worlds}), and illustrate that the set of worlds they define reduces to the three atomic worlds we identify in \autoref{sec: atomic}. We note that these RL environments are diverse in their state and action spaces; more interestingly, they are diverse in how they map to real-life tasks. Thus, we expect that many useful mHealth applications can be modeled by known atomic worlds or straightforward combinations of atomic worlds (see \autoref{sec:discussion} and \autoref{app:world_composition} for more details), allowing us to transfer intervention design from familiar, simpler settings onto unexplored and more complex ones.

Under our definition, Chain (\cref{fig:chain_behavior_map}), RiverSwim (\cref{fig:riverswim_behavior_map}), Gambler's Fallacy V1 (\cref{fig:gambler_fc_behavior_map}), and the Café worlds (\cref{fig:cafe_behavior_map}) are equivalent to the Big-Small world (\cref{fig:big_small_behavior_map}); these are worlds in which the user chooses between a readily available but small reward (i.e., disengaging in Chain, swimming downstream in RiverSwim, and performing the \emph{Finish} action in the Gambler's Ruin world) and a greater but more time-consuming reward. Gambler's Fallacy V2 (\cref{fig:gambler_ff_behavior_map}) is equivalent to Cliff World---both worlds have a ``catastrophic absorbing state," i.e., a nonzero risk of ending up in a terminal state with a negative reward.

\subsection{The Equivalence Definition Is Robust to Parameter Perturbations in World Definitions}

We want a world to remain within its equivalence class despite minor parameter adjustments (e.g., the world for a month-long PT program should be in the same class as that for a 2-month program). This is evidence that our equivalence definition captures essential rather than incidental qualities of applications.

In \cref{fig:inclass_invariance}, we verify that the Big-Small world remains within its equivalence class despite parameter changes, such as the world's width or the ratio of the big to a small reward.
In \autoref{app:param_perturbation}, we provide additional evidence of how our equivalence classes withstand perturbations across more parameters for all $8$ worlds investigated.

\subsection{Heuristics for Intervention Transfer}
\label{sec: equivalence design}
Many real-world applications may be roughly mapped to an atomic world through domain knowledge rather than computing an explicit map $h$, as in \autoref{sec: transfer}. For example, behavior scientists can often describe the types of expected user behavior, e.g., ``how many different behaviors are there for users with very low confidence?". Absent a map $h$, we cannot transfer an intervention strategy in precise terms. However, the broader insights we obtain from studying the behavior maps of atomic worlds can be easily transferred. For example, conclusions we reach on the identifiability of user traits and the effectiveness of a particular warm-start intervention strategy (see \autoref{sec: behavior maps}) apply to all worlds within the same equivalence class.

\section{Discussion \& Future Work}
\label{sec:discussion}

\paragraph{Exhaustive World Search.}
We expect there to be many equivalence classes outside the three identified in this paper. The existence of such classes may be especially relevant when we try to capture multiple distinct aspects of an mHealth application in a single world. In future work, we intend to explore the space of possible equivalence classes more exhaustively.

\paragraph{World Compositions.}
Complex real-life scenarios are unlikely to neatly map to a singular atomic world; however, we conjecture that some worlds may fall into \textit{compositions} of atomic worlds. Some initial experiments with composite worlds indicate that the composition of the Big-Small and Cliff worlds leads to a behavior map that combines the atomic worlds' respective maps. See \autoref{app:world_composition} for examples of these experiments. This finding further supports the generality of our equivalence classes as seemingly-complicated scenarios can be broken down into atomic worlds that each capture a unique aspect of the application.

\paragraph{Other User-Intrinsic Obstacles.}
While we focus on myopia ($\gamma^\text{user}$) and confidence ($T_p^\text{user}$) in this paper, we are interested in modeling a wider range of user-intrinsic obstacles, as differences between the real and user-perceived MDP. For motivation, works like \citet{evans2016learningIgnorant}, under a different model of the user's decision-making process, capture behaviors that cannot be parameterized as combinations of $\gamma^\text{user}$ and $T_p^\text{user}$ in the Café world. This observation raises the question of whether our formal framework can capture behaviors observed under other paradigms of sequential decision-making (e.g. hyperbolic discounting, replanning, etc).

\paragraph{Real World Dynamics vs. User Perceived Dynamics.}
We note that the definition of behavior maps does not rely on the environment's true dynamics $T$ since the user's policy is computed based on their perceived dynamics $T_p^\text{user}$. In reality, if $T$ and $T_p^\text{user}$ are significantly different, it would be reasonable to assume that the user iteratively updates $T_p^\text{user}$ as they interact with the real world.

\paragraph{The Topology of Behavior Maps.}
For the set of worlds in this work, verifying that any two are equivalent reduces to matching the number of behavior changes along the edges of their behavior maps. That is, the decision boundaries of their behavior maps have no interesting topology. See \autoref{app:middle_region} for a discussion on intervention transfers between worlds whose behavior maps are topologically distinct in more nuanced ways. Future research could characterize the set of worlds for which the decision boundaries of the behavior maps are not as ``well-behaved''.

\section{Conclusion}
In this work, we propose a novel tool, the behavior map, to study the relationship between user traits and user behaviors for worlds in which the user acts as an RL agent. We define an equivalence relation between worlds based on the shapes of their corresponding behavior maps. We show that intervention strategies can be transferred between equivalent worlds. In particular, we demonstrate that many seemingly different RL environments map to one of a few equivalence classes, each represented by a simple atomic world. We further argue that many real-world applications can be mapped to atomic worlds by leveraging domain knowledge in behavioral science and psychology. Finally, we show how broad insight into intervention design for simple worlds can be lifted to complex ones in the same equivalence class.

\section*{Acknowledgements}
This material is based upon work supported by the National Science Foundation under Grant No. IIS-2107391. Any opinions, findings, conclusions, or recommendations expressed in this material are those of the author(s) and do not necessarily reflect the views of the National Science Foundation.

\cleardoublepage
\bibliography{references}
\bibliographystyle{icml2023}

\pagebreak
\appendix
\onecolumn
\section{Descriptions of Each World from the Literature}
\label{appendix: worlds}

\begin{figure}[ht]
    \centering
    \includegraphics[width=0.70\linewidth]{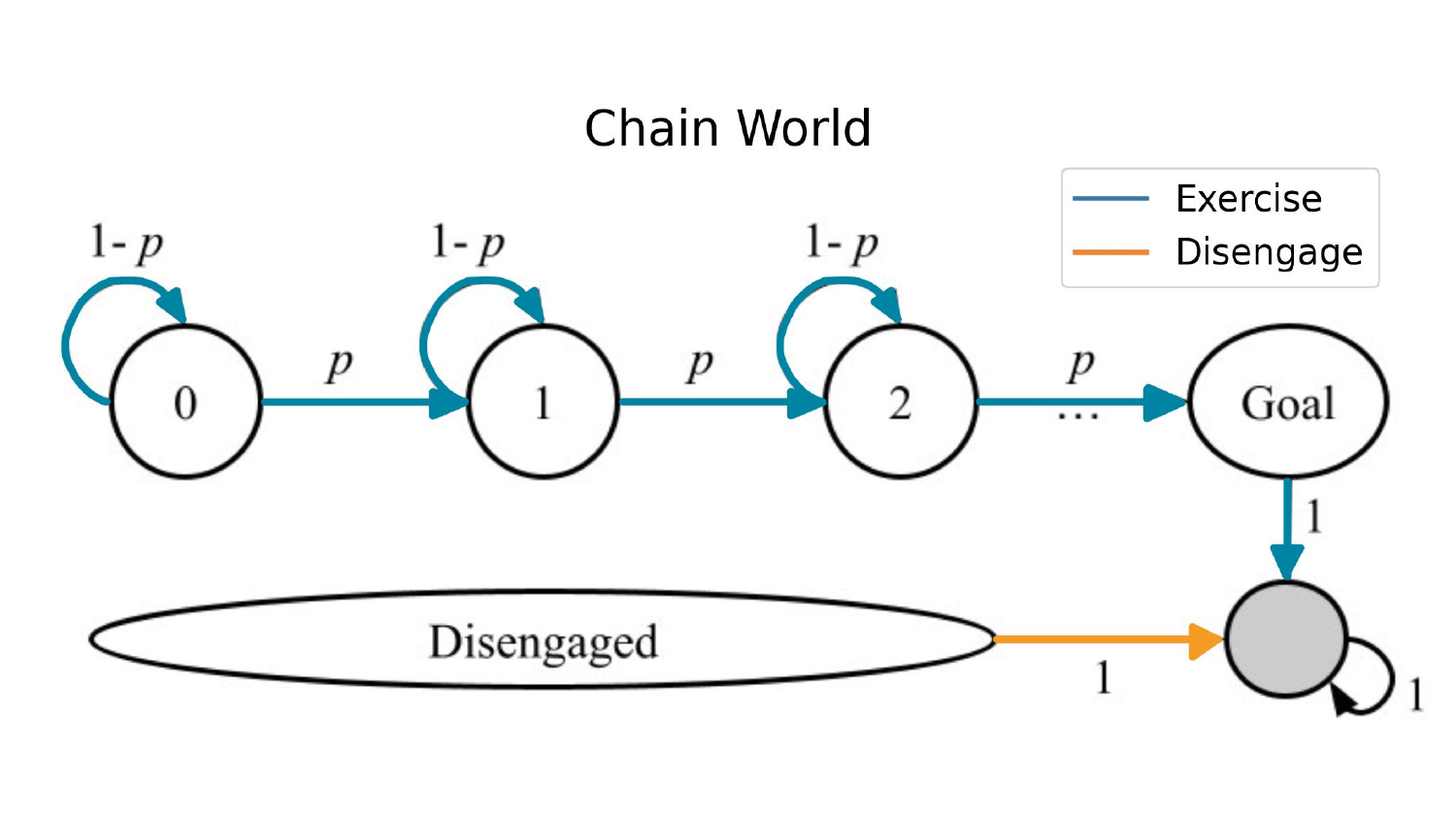}
    \caption{In the Chain world, users can choose to ``exercise,'' or progress step-by-step to reach the  desired goal. At each stage, they also have the option to ``disengage,'' which results in a smaller reward and the termination of their progression.}
    \label{fig:behaviors_chain}
\end{figure}

\begin{figure}[ht]
    \centering
    \includegraphics[width=0.9\linewidth]{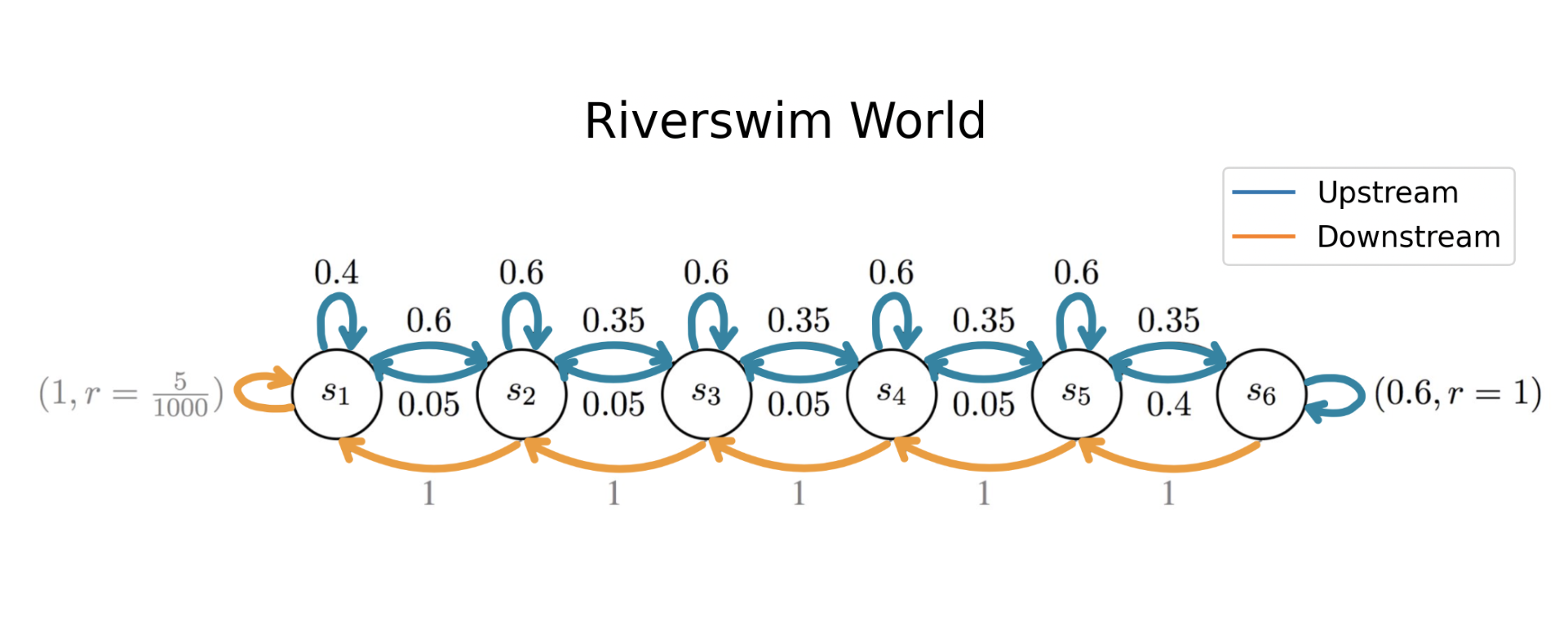}
    \caption{In the RiverSwim world, the user can choose the rightward ``upstream'' action, which has a chance of successfully advancing the user toward the larger reward but also a failure probability of staying in place or falling behind. They can also choose the leftward ``downstream'' action that deterministically moves the user toward the small reward on the far left.}
    \label{fig:behaviors_riverswim}
\end{figure}

\begin{figure}[ht]
    \centering
    \includegraphics[width=0.70\linewidth]{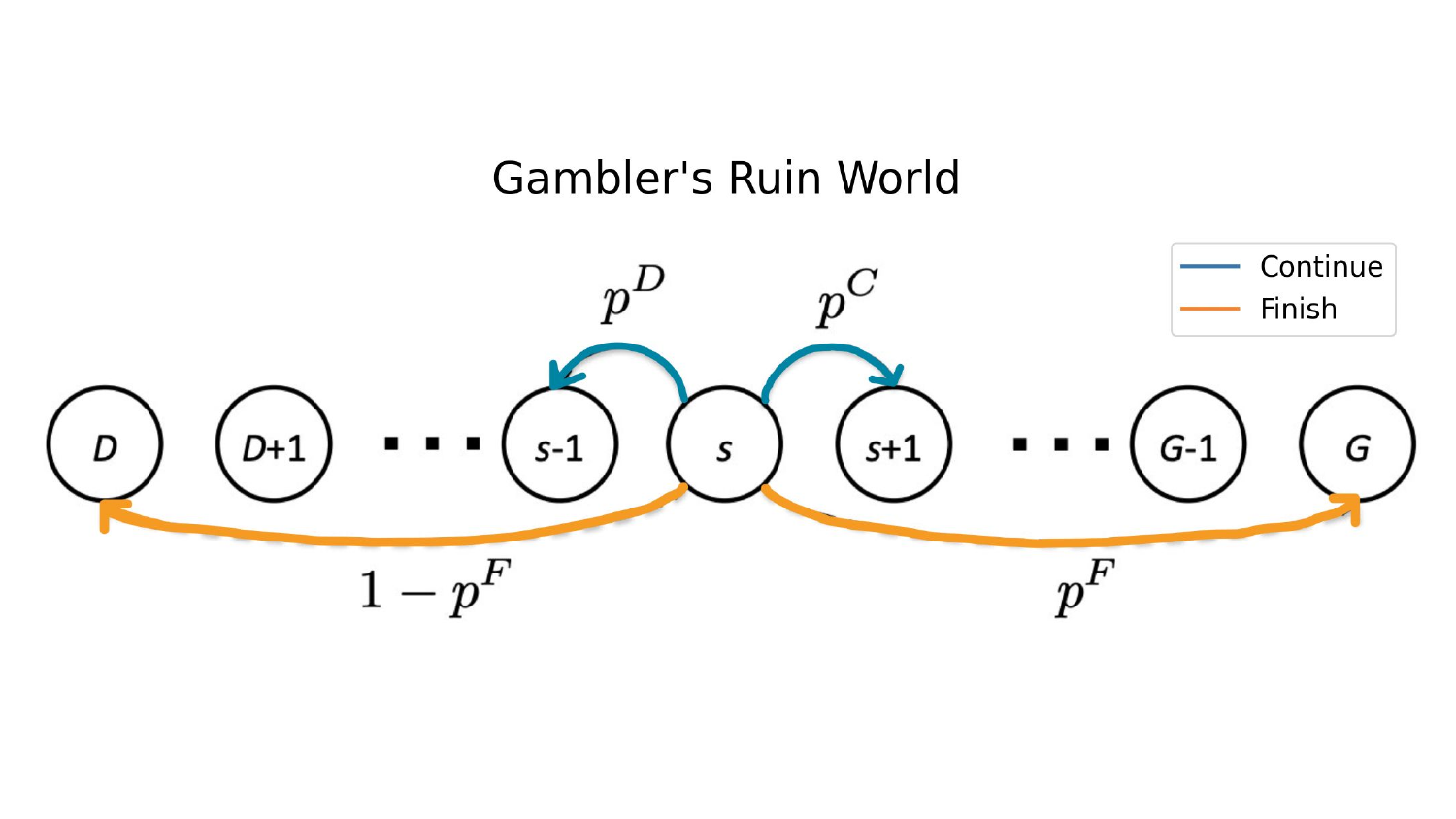}
    \caption{In the Gambler's Ruin (Bandit Problem) world, users can choose the ``continue'' action, which can either move the user one step left toward the dead-end state or one step right toward the goal state. They can also choose the ``finish'' action, moving them directly to the dead-end or goal state.}
    \label{fig:behaviors_gamblers}
\end{figure}

In this section, we present the MDPs for the MDPs from the mHealth literature we study in this work, i.e., the Chain World, RiverSwim World, and Gambler's Ruin in \cref{fig:behaviors_chain}, \cref{fig:behaviors_riverswim}, and \cref{fig:behaviors_gamblers}, respectively. Blue arrows indicate the behavior corresponding to the blue behavior in the corresponding behavior maps and, likewise, for orange arrows.

\begin{figure}[ht]
    \centering
    \includegraphics[width=0.50\linewidth]{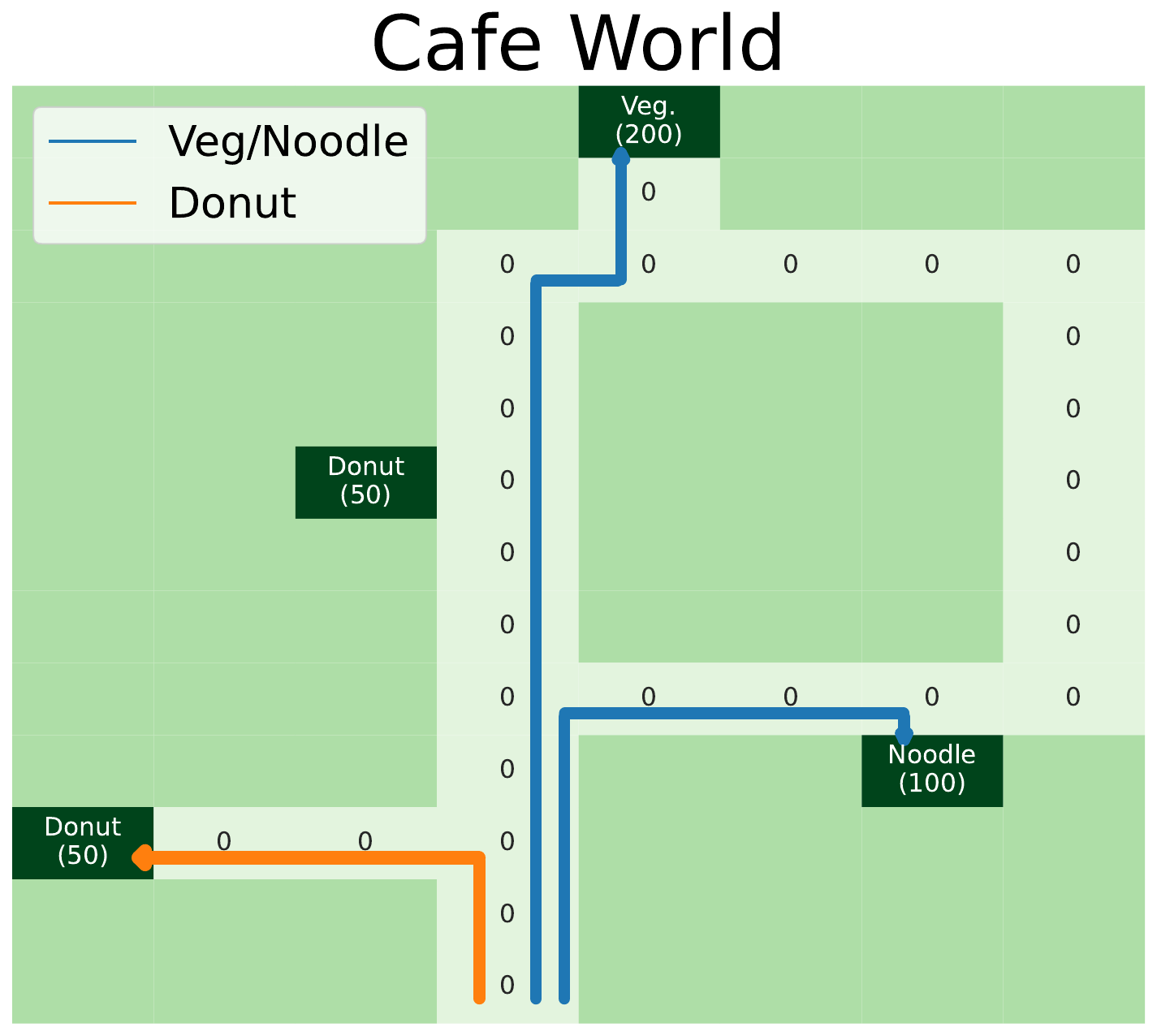}
    \caption{In the Café world, users start at the bottom of a 13 $\times$ 8 grid and must choose where to get food. The choices are two donut stores, a noodle shop, and a vegan café. The rewards of 50, 50, 100, and 200 represent the long-term satisfaction someone might feel from eating the food. An important dynamic in this world is that users must pass the donut stores to reach the noodle shop or vegan café, and the noodles are closer to the start than the vegan café. In our initial experiments, we look at the choice between the unhealthy choice (donuts) versus the comparatively healthy choice of noodles or vegan food. We indicate the paths users take when making the unhealthy choice in orange and the healthy choices in blue. In \autoref{app:world_composition}, we look at the dynamics of the behavior maps when all three choices are evaluated separately.}
    \label{fig:behaviors_cafe}
\end{figure}

\cleardoublepage
\section{Parameter Perturbations for Each World}
\label{app:param_perturbation}

In the following, we present more comprehensive investigations into the invariance of the different worlds to changes in the world parameters under our definition of equivalence. Different worlds have different sets of parameters to perturb and ranges for which they remain invariant.

\begin{figure*}[ht]
    \centering
    \includegraphics[width=0.74\textwidth]{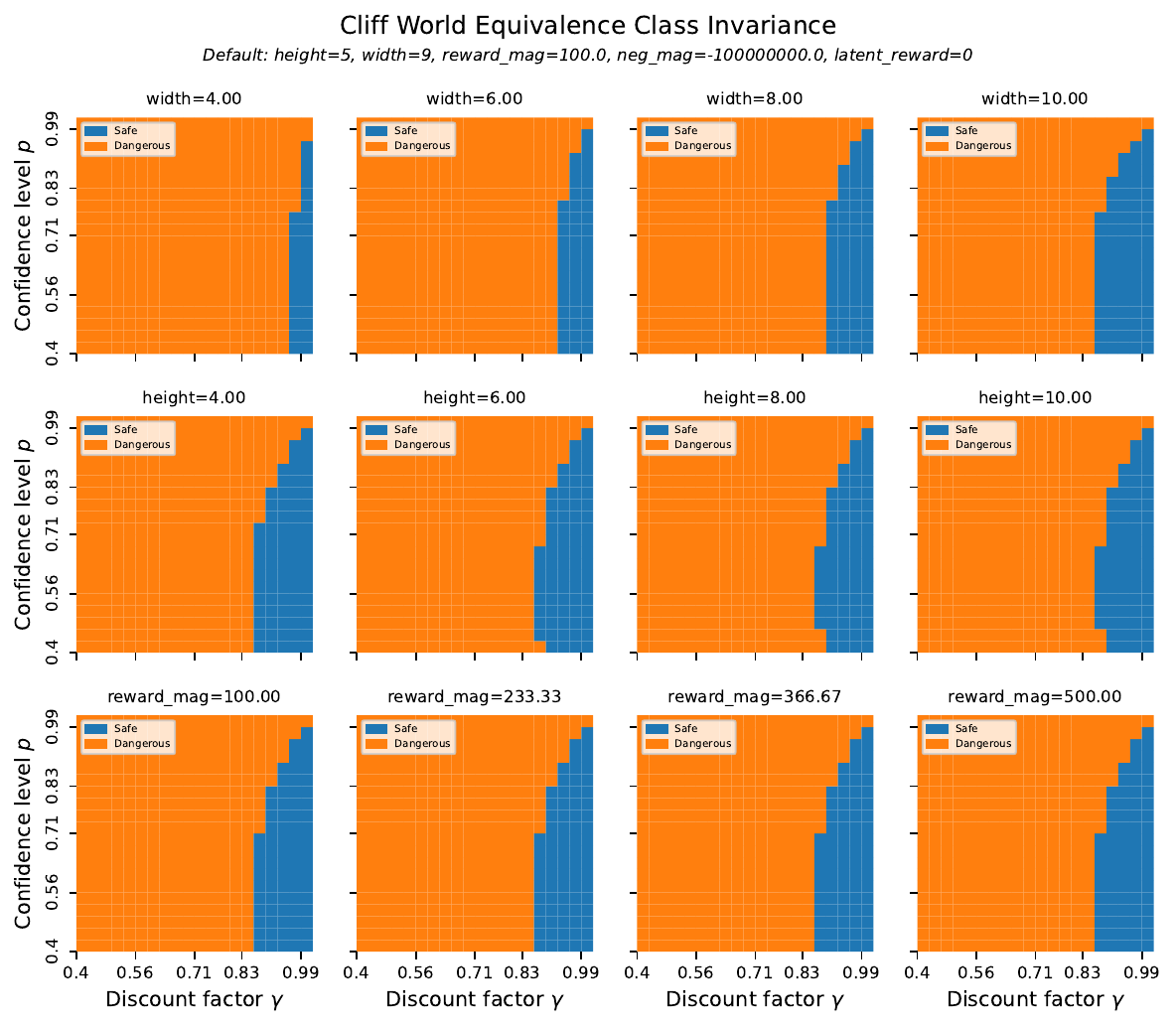}
    \caption{This array of graphs depicts behavior maps within the Cliff world across variations of three parameters: height, width, and reward size. These maps are placed in the same equivalence class under our definition, indicating their robustness to parameter perturbations.}
    \label{fig:parameter_pertubation_Cliff}
\end{figure*}

\begin{figure*}
    \centering
    \includegraphics[width=0.74\textwidth]{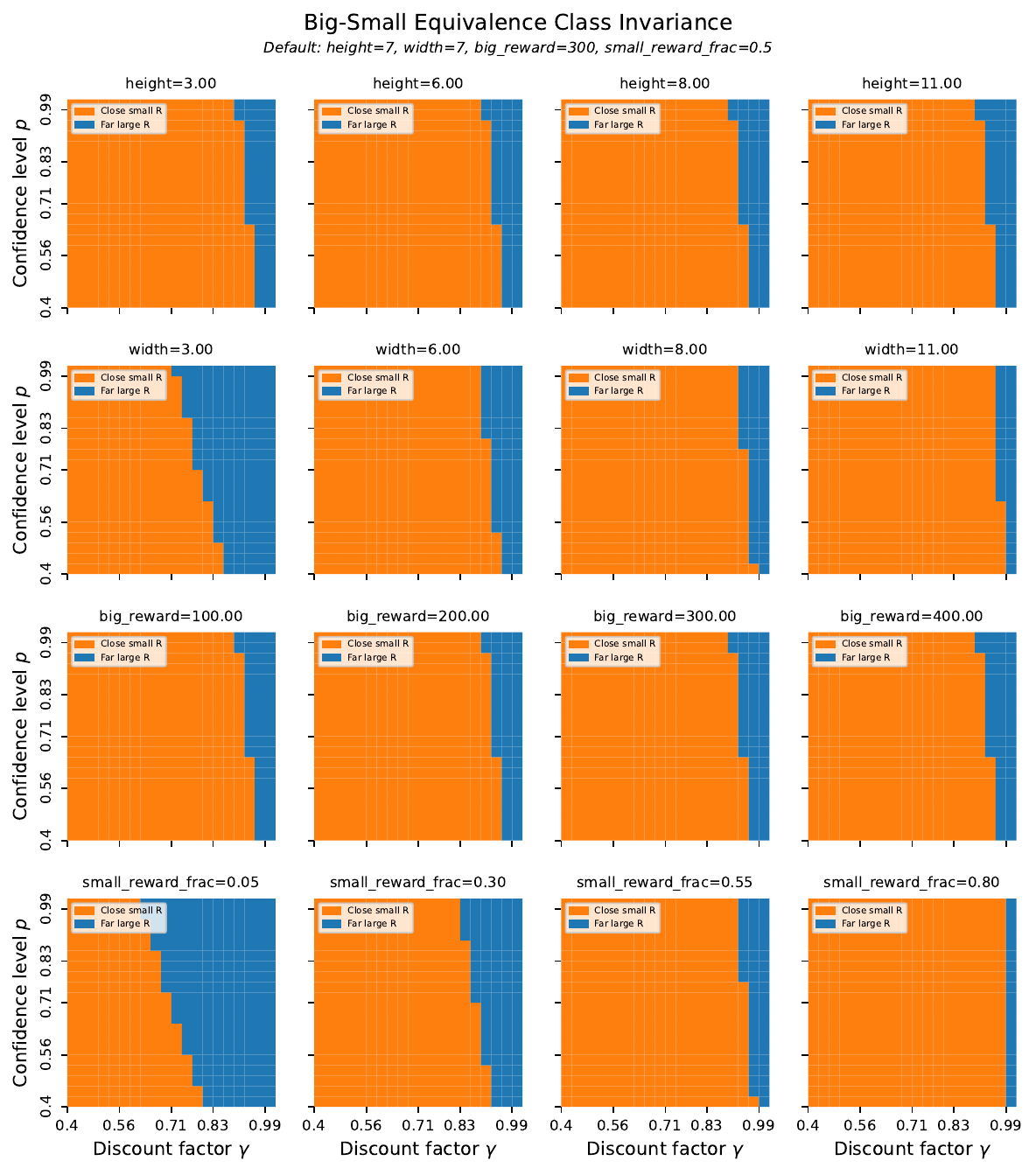}
    \caption{This array of graphs depicts behavior maps within the Big-Small world across variations of multiple parameters, such as world size and magnitude of rewards. While the graphs are not identical, all these maps are still in the same equivalence class under our definition, indicating their robustness to parameter perturbations.}
    \label{fig:parameter_pertubation_BigSmall}
\end{figure*}

\begin{figure*}
    \centering
    \includegraphics[width=0.74\textwidth]{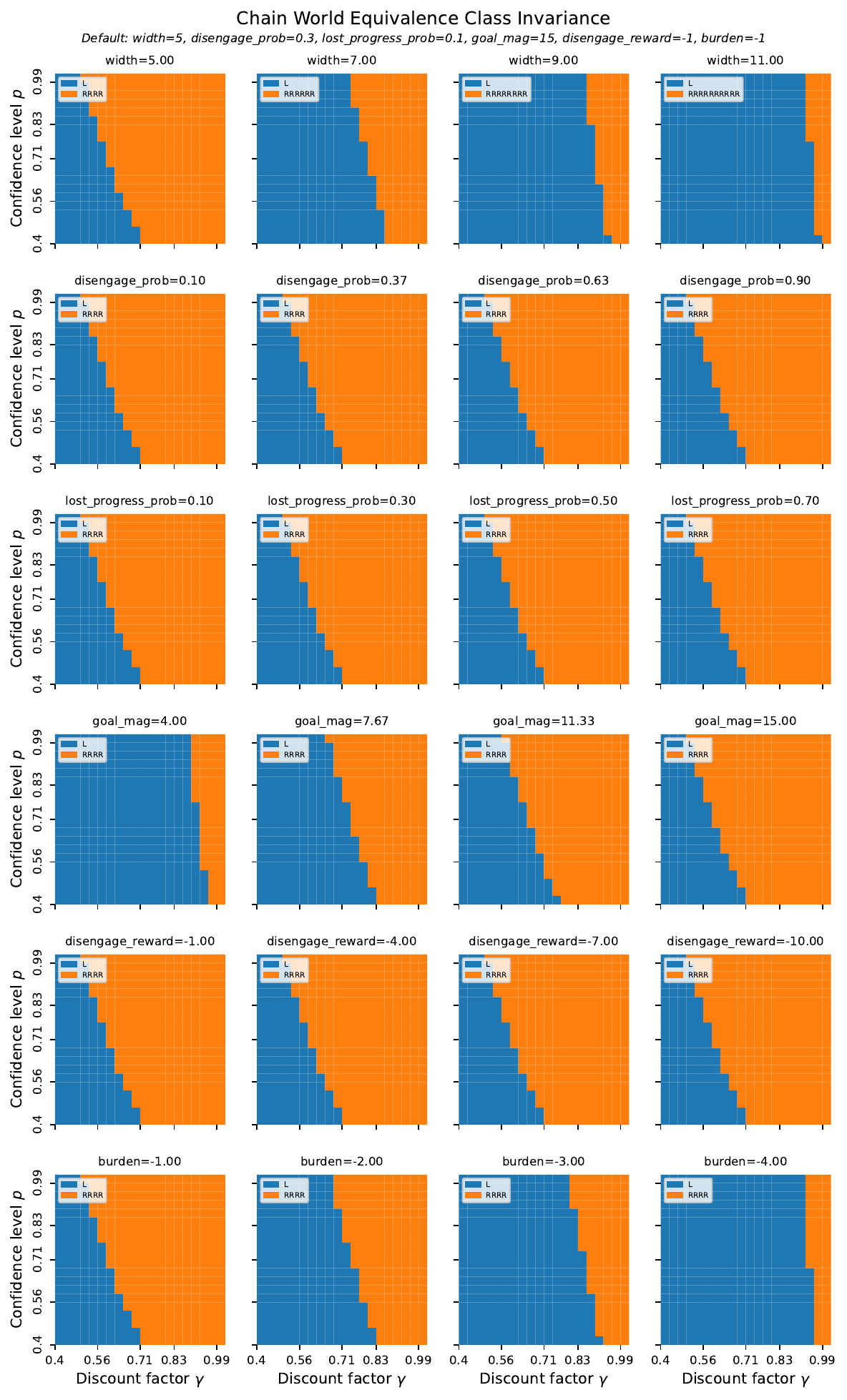}
    \caption{This array of graphs depicts behavior maps within the Chain world across variations of six parameters, including world size and disengagement probabilities. These maps are placed in the same equivalence class under our definition, indicating their robustness to parameter perturbations.}
    \label{fig:parameter_pertubation_Chain}
\end{figure*}

\begin{figure*}
    \centering
    \includegraphics[width=0.74\textwidth]{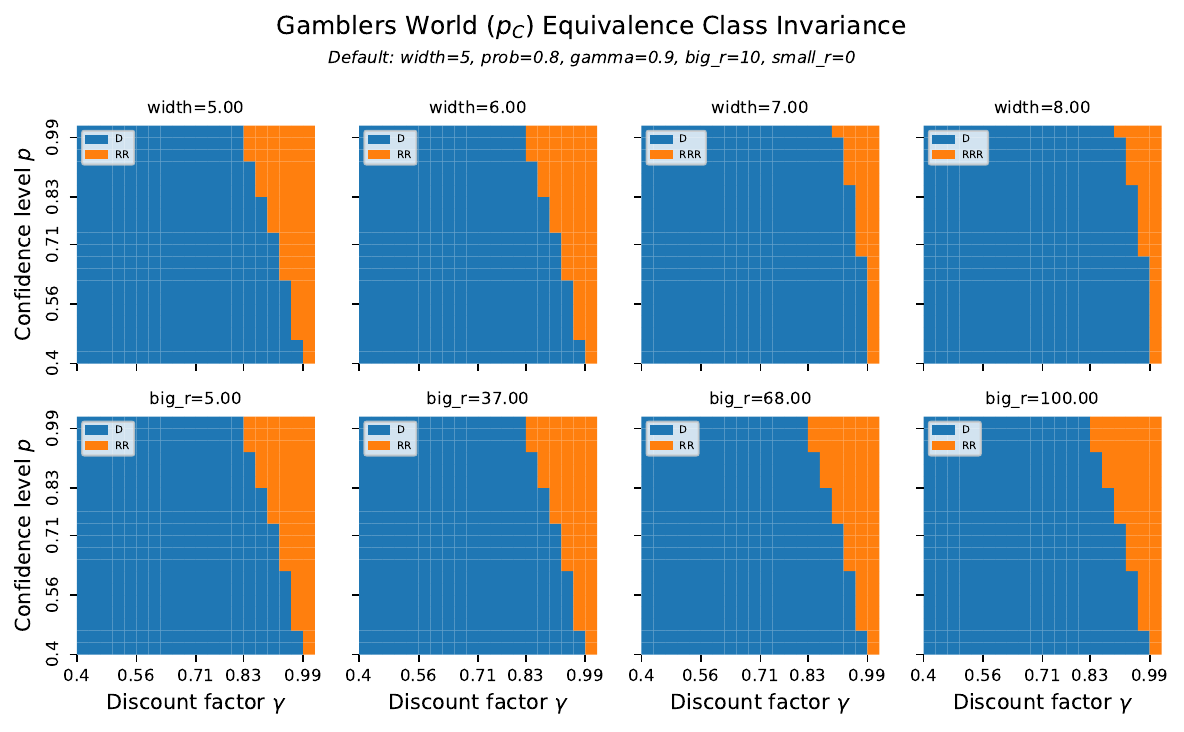}
    \caption{This array of graphs depicts behavior maps within the Gambler's Ruin world across the width and reward size variations while holding the failure probability $\left(p^F\right)$ constant. These maps are placed in the same equivalence class under our definition, indicating their robustness to parameter perturbations.}
    \label{fig:parameter_pertubation_Gamblers_pC}
\end{figure*}

\begin{figure*}
    \centering
    \includegraphics[width=0.74\textwidth]{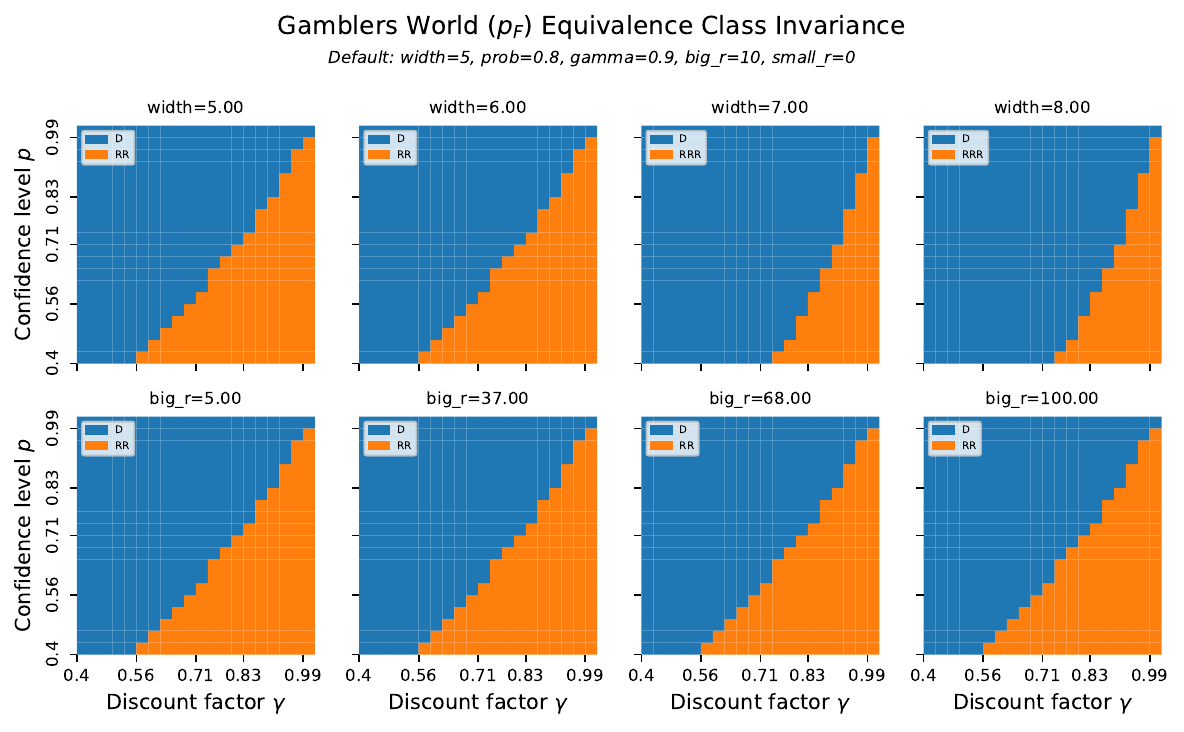}
    \caption{This array of graphs depicts behavior maps within the Gambler's Ruin world across the width and reward size variations while holding the ``continue'' probability $\left(p^C\right)$ constant. These maps are placed in the same equivalence class under our definition, indicating their robustness to parameter perturbations.}
    \label{fig:parameter_pertubation_Gamblers_pF}
\end{figure*}

\begin{figure*}
    \centering
    \includegraphics[width=0.74\textwidth]{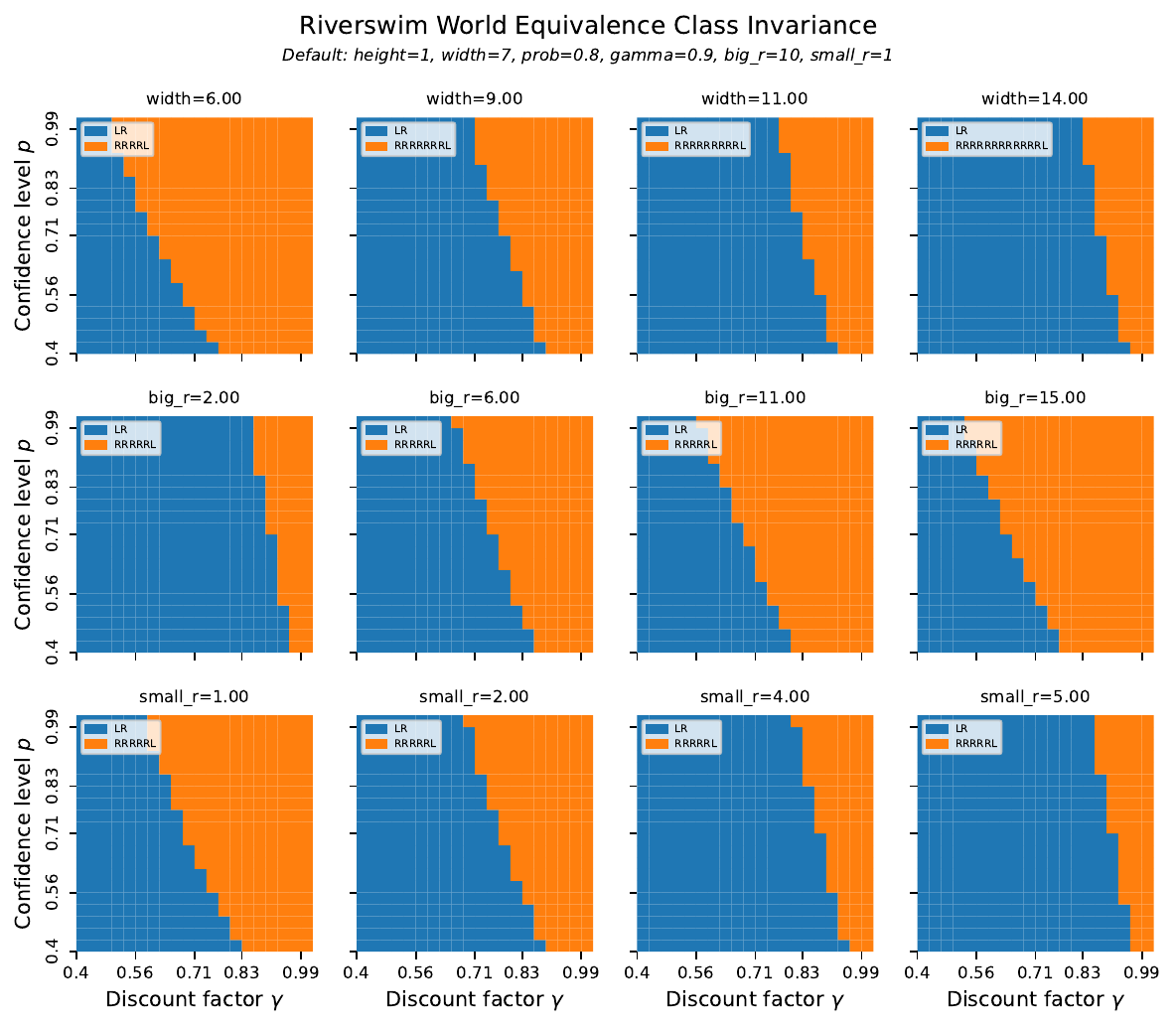}
    \caption{This array of graphs depicts behavior maps within the RiverSwim world across the width and reward sizes variations. These maps are placed in the same equivalence class under our definition, indicating their robustness to parameter perturbations.}
    \label{fig:parameter_pertubation_RiverSwim}
\end{figure*}

\begin{figure*}
    \centering
    \includegraphics[width=0.74\textwidth]{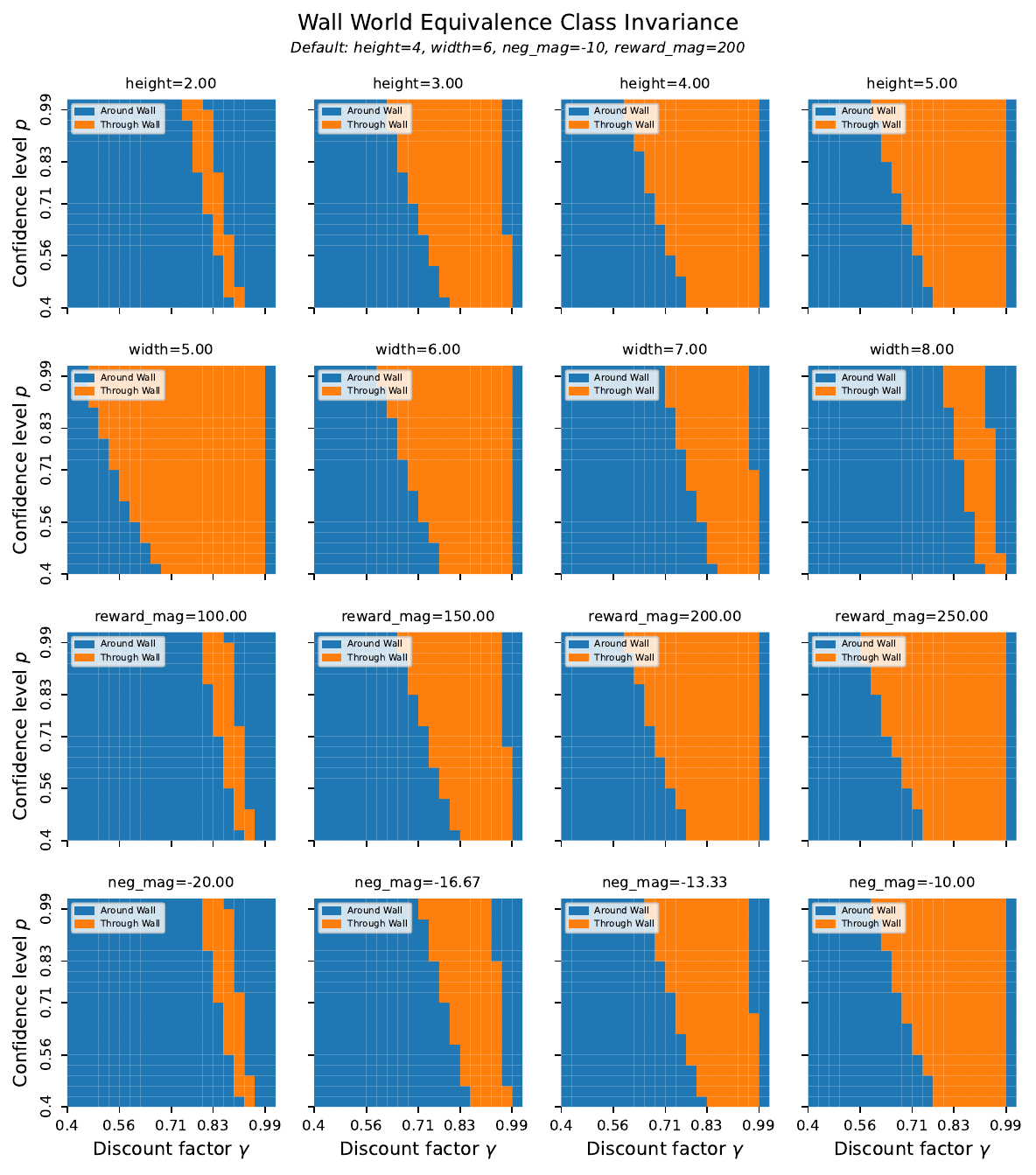}
    \caption{This array of graphs depicts behavior maps within the Wall world across variations of world size and reward magnitude. These maps are placed in the same equivalence class under our definition, indicating their robustness to parameter perturbations.}
    \label{fig:parameter_pertubation_Wall}
\end{figure*}

\begin{figure*}
    \centering
    \includegraphics[width=0.74\textwidth]{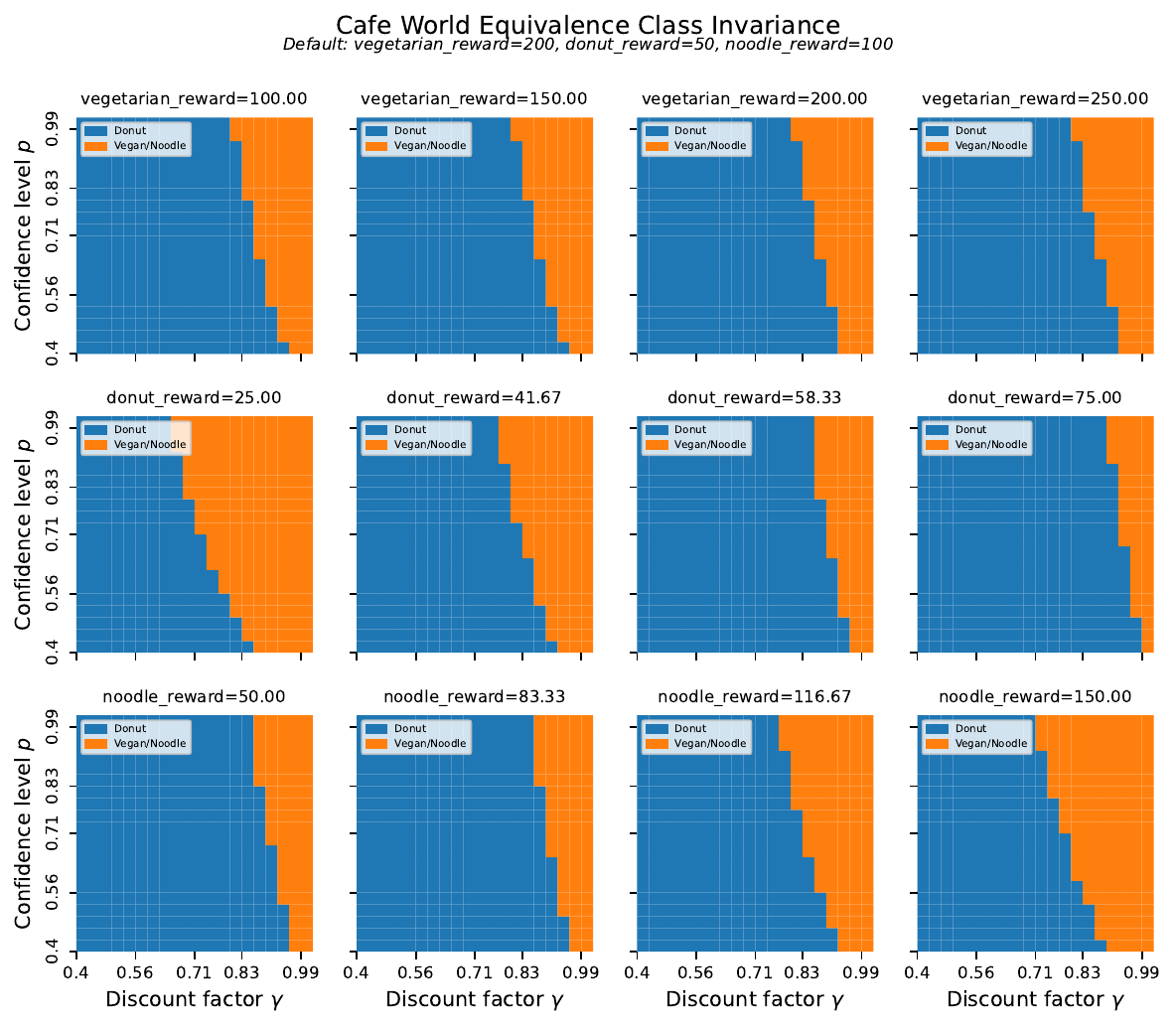}
    \caption{This array of graphs depicts behavior maps within the Cafe world, presented in \cite{evans2016learningIgnorant}, across variations of relative reward for the types of eating options: Donuts, noodles/veggie. These maps are placed in the same equivalence class under our definition, indicating their robustness to parameter perturbations.}
    \label{fig:parameter_pertubation_Cafe}
\end{figure*}

\pagebreak

\section{Initial World Composition Experiments}
\label{app:world_composition}

In this section, we present two behavior map perturbations that indicate that more complex worlds can be decomposed as a combination of several smaller atomic worlds.

\begin{figure*}[ht]
    \centering
     \begin{subfigure}[t]{0.45\linewidth}
        \centering
        \includegraphics[width=1\textwidth]{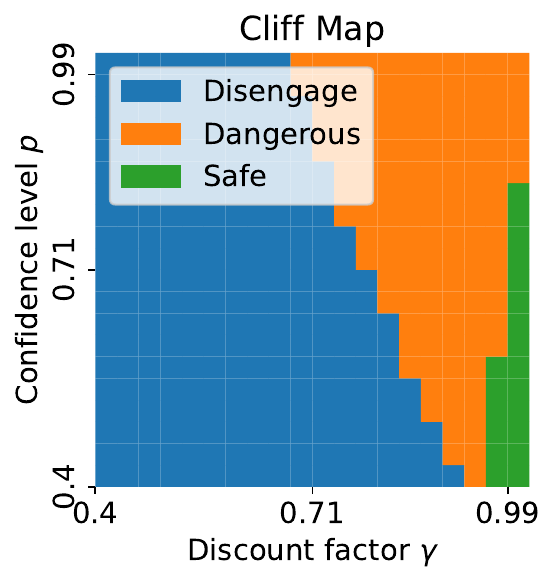}
        \caption{In the Cliff world with the possibility for disengagement, the agent is effectively faced with first the choice between a small and big reward, and then the choice of strategy for traversing the cliff (safe or risky). This effect is evident from the new decision boundary that crosses from the top edge's left side to the bottom edge's right side, just like a Big-Small decision boundary.}
        \label{fig:composition_map_Cliff}
    \end{subfigure}\hfill
    \begin{subfigure}[t]{0.45\linewidth}
        \centering
        \includegraphics[width=1\textwidth]{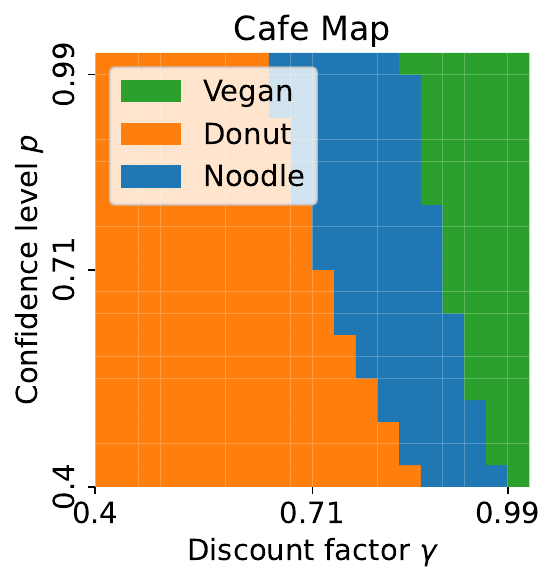}
        \label{fig:composition_map_Cafe}
        \caption{In the Café world, the agent is effectively faced with two, sequential Big-Small worlds. This can be seen by considering the boundary between the orange and blue areas as the first decision between a far and big reward (noodle/vegan) and a near and small reward (donuts). If the user avoids the donuts, they are faced with the choice between another far and big reward (vegan) and a near and (relatively) small reward (noodle).}
    \end{subfigure}
    \caption{Two worlds that appear to be straightforward compositions of two atomic worlds. Characterizing these compositions and understanding whether and how they can be useful is an interesting avenue for future research.}
    \label{fig:composition_worlds}
\end{figure*}

\paragraph{Big-Small \& Cliff Composition.} The first, seen in \cref{fig:composition_map_Cliff}, is a Cliff world with an added option of disengaging. This disengagement state is modeled as a state immediately below the start state in \cref{fig:behaviors_cliff}. The disengagement is associated with a small positive reward, which can, e.g., be interpreted as the user's sense of relief for not having to engage in physical therapy anymore (which is obviously smaller than the faraway reward of being fully rehabilitated). The compositionality comes from the observation that the user now has two choices: (1) to engage or disengage, and (2) if they engage, be safe, or take risks. The first choice is similar to a Big-Small world (disengage for a small reward or engage for an expected bigger reward farther away).

\paragraph{Big-Small \& Big-Small Composition.} The second composition, whose behavior map is shown in \cref{fig:composition_map_Cafe}, is the Café world with the choices between donuts, noodles, and vegan. Intuitively, the agent is now faced with two separate decisions, where both are the choice between a small reward near and a relatively larger reward farther away.

\pagebreak

\section{Consideration on the Interior of Behavior Maps}
\label{app:middle_region}

\begin{figure*}[ht]
    \centering
    \includegraphics[width=0.6\textwidth]{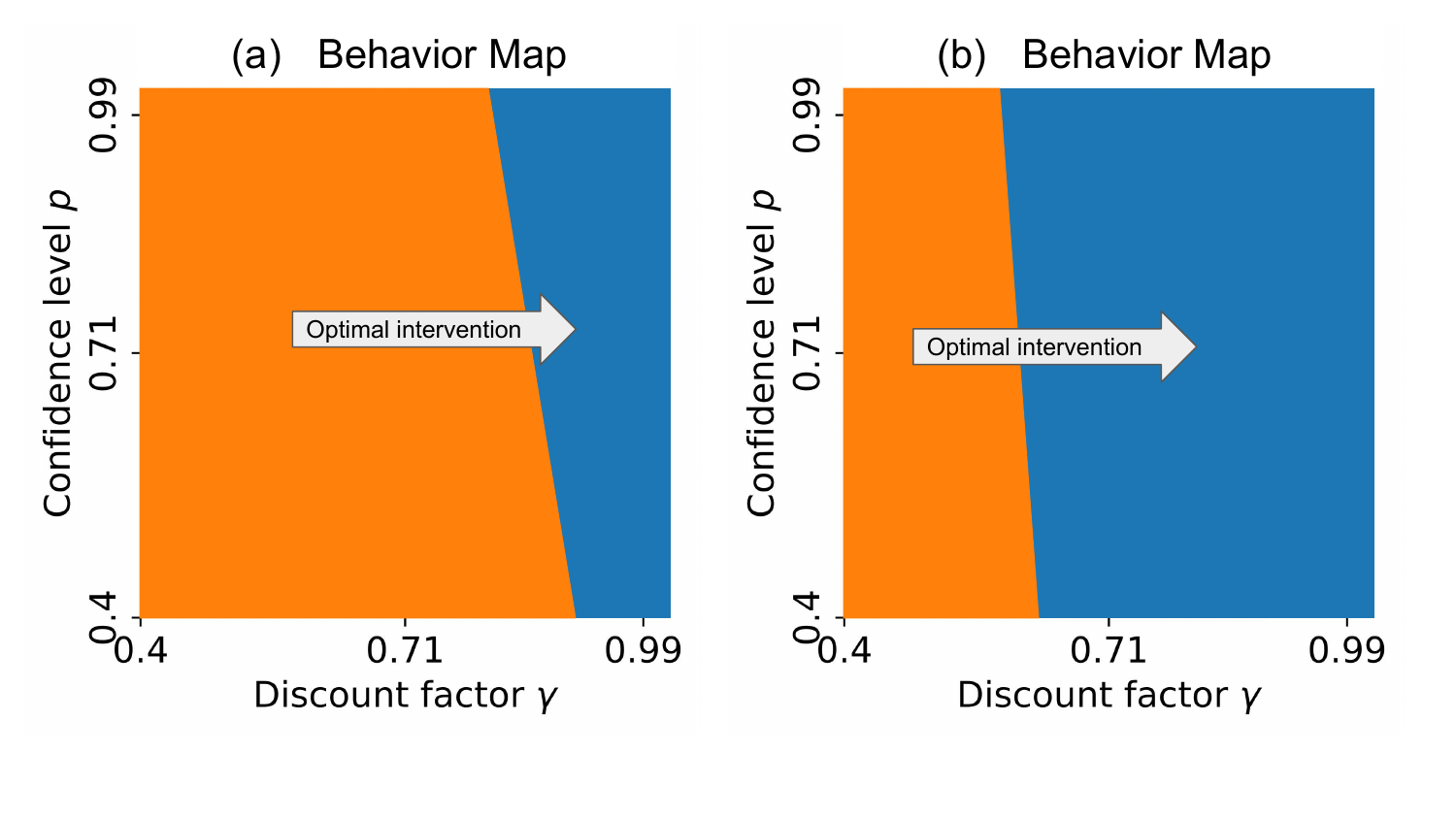}
    \caption{Two different worlds with equivalent and simple behavior maps. Gray arrows indicate the optimal intervention for an agent that exists in the orange region. Despite having their decision boundaries in different locations along the $\gamma$-axis, the best intervention is the same.}
    \label{fig:behavior_map_middle_1}
\end{figure*}

We have argued that the most important part of behavior maps is the extreme regions, i.e. the behavior along the edges. One way to argue this is by using literature from the behavioral sciences, which has been one motivating factor. Another observation one can make is the following. Let worlds (a) and (b) in \cref{fig:behavior_map_middle_1} be two different worlds that belong to the same equivalence class [1, 0, 1, 0]. Since $n_1=n_3=1$, there can be no ambiguity about how the vertices are connected. However, we place no restrictions on where along the $\gamma$-axis the vertices are. If we decide the blue region is the desired behavior, but we observe the user in the orange region (regardless of where), the optimal intervention will be the same in both worlds (a) and (b).

\begin{figure*}[ht]
    \centering
    \includegraphics[width=0.6\textwidth]{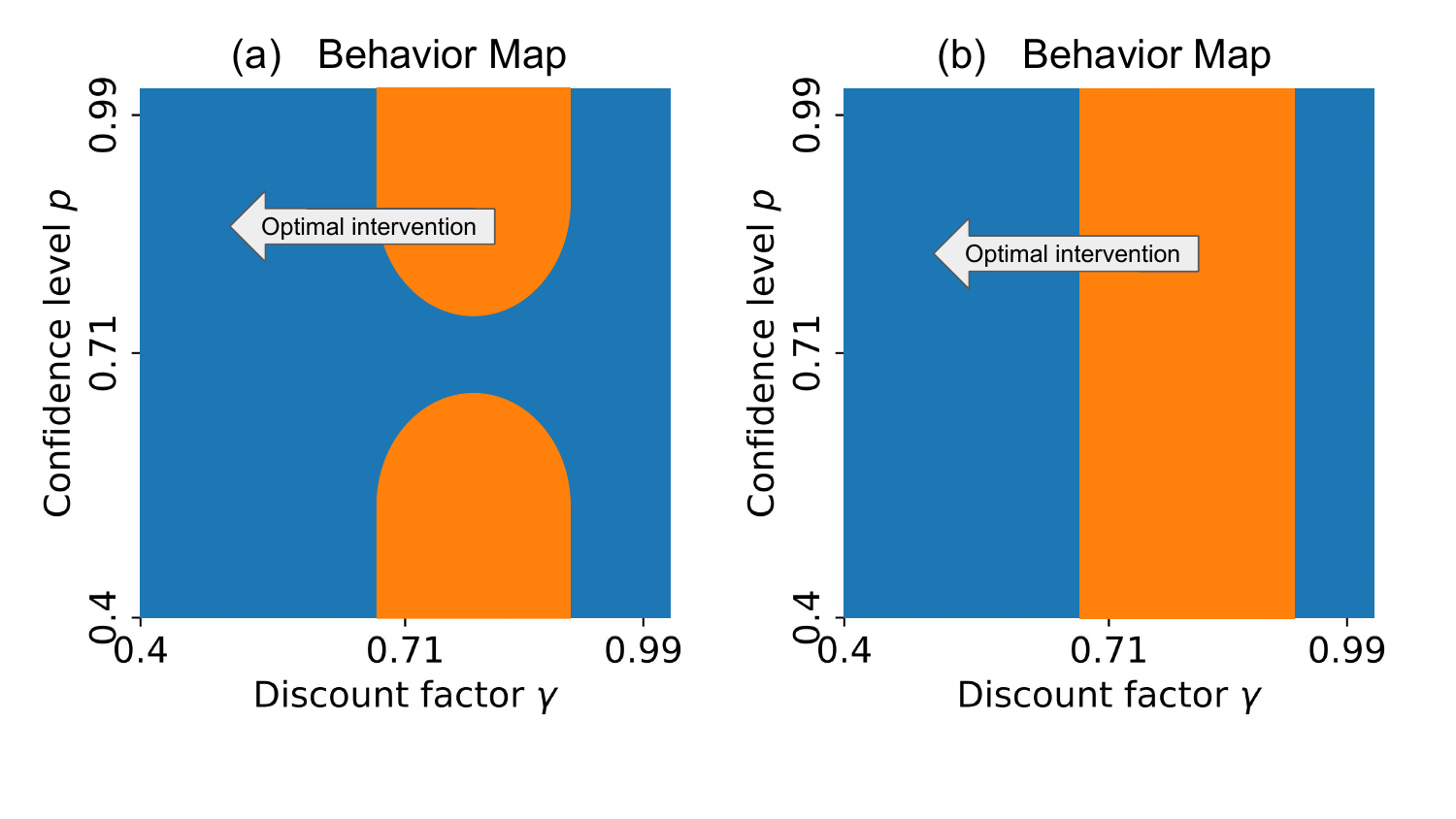}
    \caption{Two different worlds with more complex and differing behavior maps still belong to the same equivalence class. Despite having very different interior regions, in many cases, the optimal intervention on an agent located in the orange region would be the same.}
    \label{fig:behavior_map_middle_2}
\end{figure*}

In the more complex case shown in \cref{fig:behavior_map_middle_2}, both worlds (a) and (b) are in the same equivalence class [2, 0, 2, 0], despite having very different middle regions. This disparity arises since $\sum_i n_i\geq 4$ and there will be more than one valid way to connect the vertices. If we again imagine that blue is the desired behavior and orange is observed, these two worlds will still share the optimal behavior, as indicated with gray arrows in \cref{fig:behavior_map_middle_2}.

We have not proven this exhaustively; atomic worlds where this observation does not hold might still arise. From our initial experiments, however, worlds with $\sum_i n_i \geq 4$ appear rare.

\end{document}